\documentclass[conference]{IEEEtran}
\IEEEoverridecommandlockouts

\usepackage{cite}
\usepackage{amsmath,amssymb,amsfonts}
\usepackage{algorithmic}
\usepackage{graphicx}
\usepackage{textcomp}
\usepackage{xcolor}
\def\BibTeX{{\rm B\kern-.05em{\sc i\kern-.025em b}\kern-.08em
    T\kern-.1667em\lower.7ex\hbox{E}\kern-.125emX}}
\usepackage{hyperref}
\hypersetup{
  pdftitle={Optimizing BioTac Simulation for Realistic Tactile Perception},
  pdfsubject={Robotics (cs.RO); Artificial Intelligence (cs.AI),},%
  pdfauthor={Wadhah Zai El Amri and Nicolás Navarro-Guerrero},
  pdfkeywords={BioTac, XGBoost, Transformer, Tactile Perception, Simulation},%
  colorlinks=true,
  linktoc=all,
  linkcolor=darkgray,
  citecolor=darkgray,
  urlcolor=black,
}
\usepackage[protrusion=true,expansion=true]{microtype}
\usepackage{textcomp}
\usepackage{multirow}
\usepackage{blindtext}
\usepackage{footmisc}
\usepackage{subfigure}
\usepackage{xcolor}

\usepackage{fancyhdr}
\fancypagestyle{FirstPage}{
\lfoot{\footnotesize{\copyright 2024 IEEE.  Personal use of this material is permitted.  Permission from IEEE must be obtained for all other uses, in any current or future media, including reprinting/republishing this material for advertising or promotional purposes, creating new collective works, for resale or redistribution to servers or lists, or reuse of any copyrighted component of this work in other works.} }
}
\begin{document}

\title{Optimizing BioTac Simulation for Realistic Tactile Perception}


\author{\IEEEauthorblockN{Wadhah Zai El Amri}
\IEEEauthorblockA{\textit{L3S Research Center, Leibniz Universität Hannover} \\
Hannover, Germany \\
\href{https://orcid.org/0000-0002-0238-4437}{0000-0002-0238-4437}}
\and
\IEEEauthorblockN{Nicolás Navarro-Guerrero}
\IEEEauthorblockA{\textit{L3S Research Center, Leibniz Universität Hannover} \\
Hannover, Germany \\
\href{https://orcid.org/0000-0003-1164-5579}{0000-0003-1164-5579}}}

\maketitle

\begin{abstract}
Tactile sensing presents a promising opportunity for enhancing the interaction capabilities of today's robots. BioTac is a commonly used tactile sensor that enables robots to perceive and respond to physical tactile stimuli. However, the sensor's non-linearity poses challenges in simulating its behavior. 
In this paper, we first investigate a BioTac simulation that uses temperature, force, and contact point positions to predict the sensor outputs. We show that training with BioTac temperature readings does not yield accurate sensor output predictions during deployment. Consequently, we tested three alternative models, i.e., an XGBoost regressor, a neural network, and a transformer encoder. We train these models without temperature readings and provide a detailed investigation of the window size of the input vectors. 
We demonstrate that we achieve statistically significant improvements over the baseline network.
Furthermore, our results reveal that the XGBoost regressor and transformer outperform traditional feed-forward neural networks in this task.
We make all our code and results available online on \href{https://github.com/wzaielamri/Optimizing_BioTac_Simulation}{https://github.com/wzaielamri/Optimizing\_BioTac\_Simulation}.
\end{abstract}


\section{Introduction}
\thispagestyle{FirstPage}
Tactile sensing sensors offer robots valuable information that can be used to enhance and complement knowledge coming from other modalities such as vision or audio, especially in situations where this knowledge is entirely or partially not available~\cite{Navarro-Guerrero2023VisuoHaptic, parmiggiani_20_tactile}.

In situations where the sensor is unavailable or experiment repetitions are costly, the value of a reliable, real-time simulation becomes evident. Such a simulation can effectively estimate sensor outputs for various touch scenarios. 
Such simulation would offer a good alternative to gathering data in different setups and environments \cite{Navarro-Guerrero2023VisuoHaptic}. However, simulations only approximate real-world data, and models trained on this data alone can typically not be directly used in real robots. This problem is also known as the reality gap, and it can be mitigated with techniques such as sim2real~\cite{Josifovski2022Analysis} or continual learning~\cite{Auddy2023ContinualLearning}.
Nevertheless, the quality of the simulated data should be as high as possible to minimize the reality gap and increase the success of other modules down the pipeline. 

Various studies have presented simulations for a range of tactile sensors, including TACTO~\cite{Wang2020TACTO} for vision-based sensors like DIGIT~\cite{Lambeta2020DIGIT} and OmniTact~\cite{Padmanabha2020OmniTact}, touch simulation for fabric-based tactile sensors~\cite{Melnik2021Sim}, a simulation for the iCub skin~\cite{Geukes2017Github}, and several simulations for the BioTac sensor~\cite{Ruppel2018,Narang2021,Zapata2021}. 

However, in this paper, we focus on the BioTac~\cite{Wettels2007,Wettels2014}, one of the most widely used tactile sensors~\cite{Natale2016}. It consists of a rigid core covered by an elastomeric skin filled with an incompressible conductive fluid. This sensor proved to be helpful in different tasks, such as grasp stability~\cite{Chebotar2016}, object identification~\cite{Xu2013}, localizing artificial tumors~\cite{Arian2014,Pacchierotti2016}, etc.

Some simulations of the BioTac sensor exist. Ruppel et al.~\cite{Ruppel2018} collected a real-world dataset with the BioTac 2P sensor~\cite{BioTac2PDatasheet} mounted on a Shadow Robot Hand~\cite{ShadowHand}. The dataset captures the tactile sensor readings while touching with different forces, positions, and orientations, an indenter with a spherical tip of radius equal to $2\;mm$, attached to a calibrated force-torque sensor (ATi nano17e~\cite{atiNano17}). They propose a neural network model that can estimate electrodes, vibration, and pressure signals using the force and position vector of the contact point alongside temperature values. However, their approach requires using the temperature values in the input vector, which are not available in rigid body simulation environments like Gazebo~\cite{Koenig2004gazebo}. 

Narang et al.~\cite{Narang2021} developed a finite element model (FEM) that can mimic the deformation behavior of the elastomeric skin and the liquid gel inside. From the point cloud of the FEM nodal field data, a neural network based on the PointNet++ architecture~\cite{qi2017pointnet} is used to estimate the electrode values of the BioTac 2P~\cite{BioTac2PDatasheet}. To validate the FEM model and train the neural network, they collected data on different trajectories using nine different indenters interacting with three different tactile sensors. The FE simulation takes 7 minutes to simulate one single trajectory, which makes it a big drawback. Later, they improved the approach using the NVIDIA Isaac Gym simulator~\cite{nvidiaIsaacGym}, achieving a 75 times improvement \cite{narang2021isaac}. However, a single trajectory simulation of $6\;mm$ takes $5.57\;s$ to simulate. 

Furthermore, the authors improved the electrodes' estimation, by using a self-supervised latent learning network that maps between the FEM Mesh and electrode signals. Despite these improvements, this solution is still not fast enough to be used in real-time systems. Moreover, this solution estimates only the electrodes' values and not the pressure or vibration signals recorded by the BioTac sensor, which are valuable modalities when used in grasping and manipulation tasks~\cite{Taunyazov2021}. 

Zaoata-Impata and Gil~\cite{Zapata2021} used vision to estimate the electrode values of the BioTac SP sensor~\cite{BioTacSPDatasheet}. A modified Semi-Regression Generative Adversarial Network (SR-GAN)~\cite{Olmschenk2019} was used to synthesize electrode values out of point cloud representation of grasped deformable daily objects and their grasping data by using labeled and unlabeled input data. The authors collected 4000 samples of grasping positions, electrodes, and pressure values of two BioTac SP sensors mounted on a Shadow Robot Hand~\cite{ShadowHand} and 3D point clouds recorded with an Intel RealSense D415 depth camera. Since collecting data is not trivial, a generator was used to generate synthetic BioTac data for the point clouds and the grasping positions in order to augment the dataset and improve the learning of the discriminator, whose task was to estimate the correct BioTac outputs. The unlabeled samples boosted the performance of the network. In addition, this work sheds light on the possibility of using vision, in this case through point clouds, to estimate signal data for the BioTac sensor. However, this implementation only used 4000 samples.


Since vibration and pressure readings add valuable information when used for manipulation tasks~\cite{Taunyazov2021}, we must simulate these alongside electrode values. The solution of Narang et al.~\cite{narang2021isaac} only estimates the electrode signals and is comparatively slow. In comparison, Zaoata-Impata and Gil~\cite{Zapata2021} implementation uses a small dataset. Considering these factors, we opt for the solution provided by Ruppel et al.~\cite{Ruppel2018}. 

In the next Section~\ref{sec:biotacSim}, we revisit the research conducted by Ruppel et al.~\cite{Ruppel2018}. Within that section, we first delve into the details of their work. Subsequently, in Section~\ref{sec:Methodology}, we thoroughly investigate two directions to improve the simulation further.

\section{BioTac Sensor Simulation}
\label{sec:biotacSim}
\begin{figure}[htbp]
\hfill
\subfigure[]{\includegraphics[width=0.25\columnwidth]{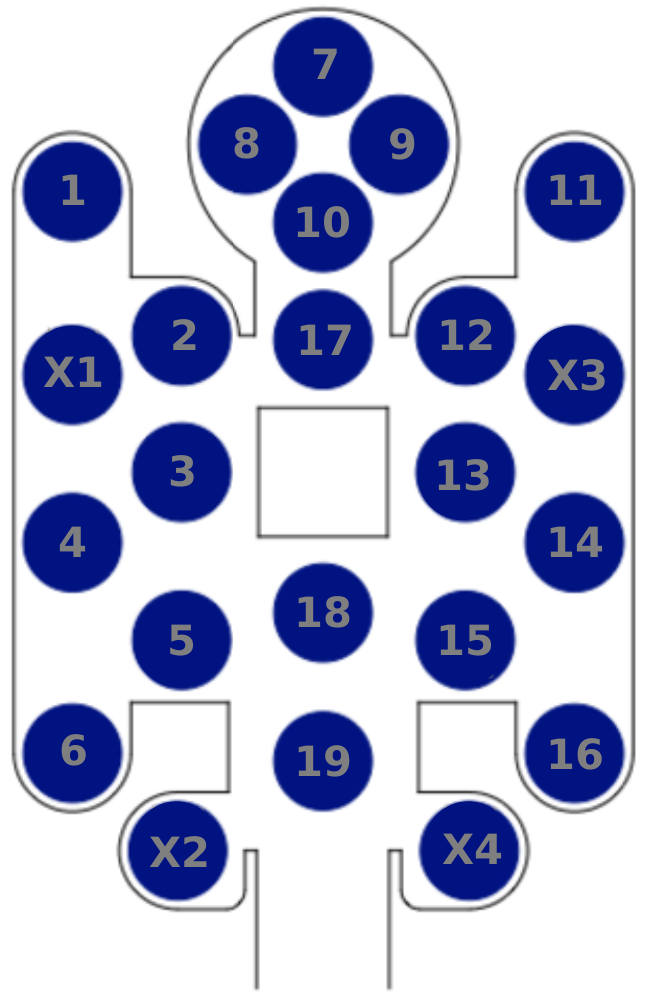} \label{fig:biotacFrame_a}}
\hfill
\subfigure[]{\includegraphics[width=0.6\columnwidth]{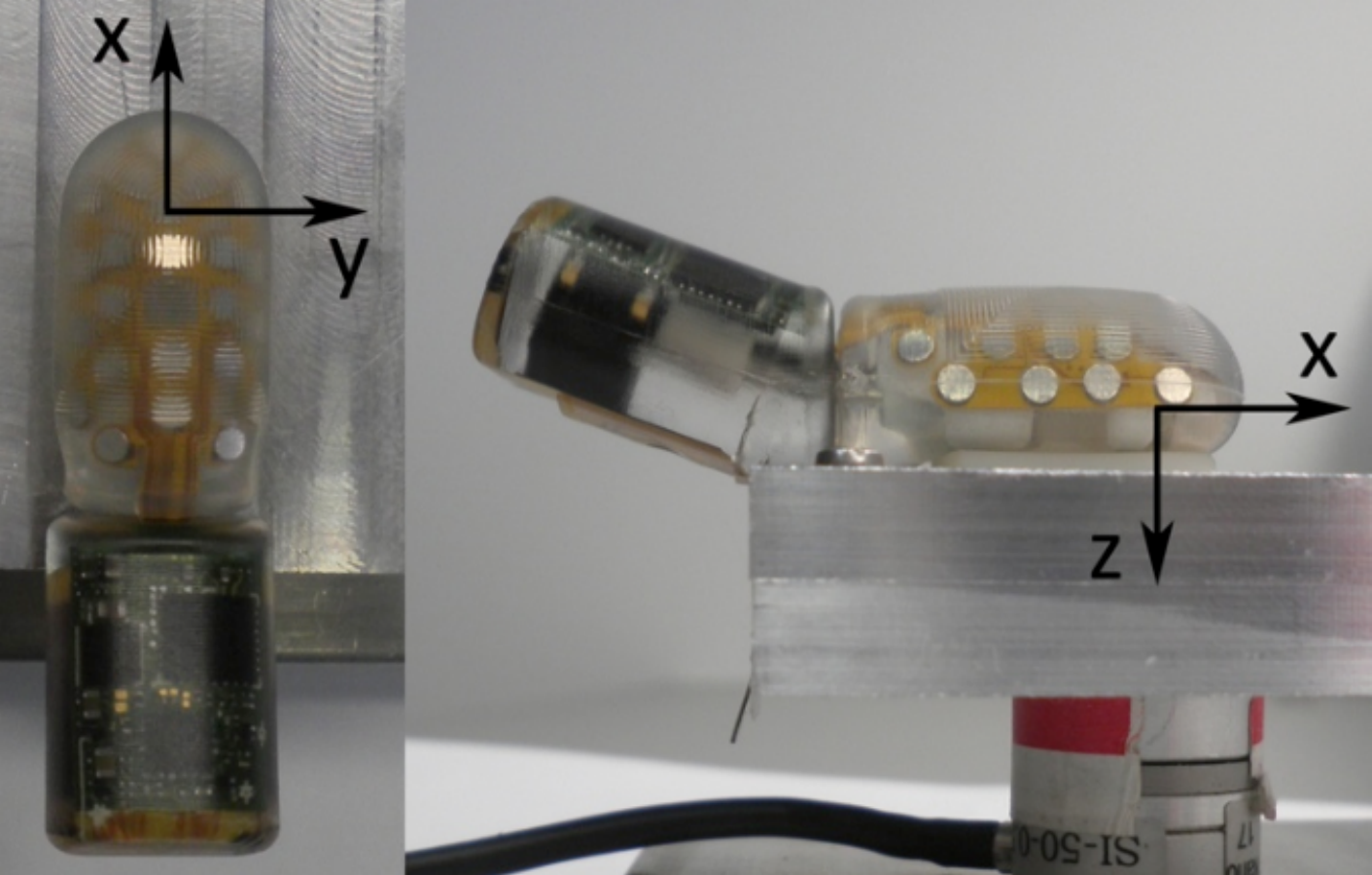}\label{fig:biotacFrame_b}}
\hfill
\caption{(a):  Map of the electrodes on the BioTac sensor~\cite{Lin2013}. (b): BioTac sensor and the position/ orientation of the coordinate frame~\cite{Lin2013}.}
\label{fig:biotacFrame}
\end{figure}

The simulation provided by Ruppel et al.~\cite{Ruppel2018} is based on a neural network trained with a real-world dataset to predict the BioTac sensor signals. This dataset includes data from approximately one hour of recording sampled at 100 Hz. It consists of the complete BioTac output values, i.e., 19 electrode voltage \textit{$e_{1}$, …, $e_{19}$}, absolute and dynamic fluid pressure \textit{pdc} and \textit{pac}, temperature \textit{tdc} and heat flow \textit{tac}. The dataset also includes the BioTac's and indenter's position and orientation recorded with a vision tracking system~\cite{Olson2011AprilTag}. 
For the data pre-processing pipeline, the position of the indenter's tip is calculated using transformation matrices in the BioTac coordinate frame, represented in Figure~\ref{fig:biotacFrame}.

These transformation matrices are calibrated to adjust for systematic offsets by reducing the error between the probes' positions and the BioTac surface. In other words, probes with light touch contact at the end of a contact cycle and with force measurements below $0.3;N$ are gathered, and their distance to the skin surface of the BioTac is calculated with the transformation matrices. In such cases, that distance should be equal to 0. Due to calibration issues, this distance deviates from 0, $5.039\;mm$ on average. While iterating for 1000 steps and adjusting these transformation matrices, i.e., adding and subtracting small values, the distance error is reduced to $0.444\;mm$ on average for all these selected probes.

Two deep neural networks, $A$ and $B$, were proposed to simulate the sensor outputs at timestep $t=T$. Both networks take as input the $x$, $y$, and $z$ contact position of the indenter's tip in the BioTac coordinate frame at timestep $t=T$, $F_x$, $F_y$, and $F_z$ values of the force sensor at timesteps $T$, $T-10$, and $T+10$ and the temperature directly taken from BioTac readings at timestep $t=T$, i.e., an input vector of length 13.

Network $A$ is a dense neural network with five hidden layers and over 6 million parameters, employing a pre-mapping fusion strategy~\cite{Navarro-Guerrero2023VisuoHaptic}. On the other hand, Network $B$ is a more compact neural network with around 800 thousand parameters and adopting a midst-mapping fusion strategy~\cite{Navarro-Guerrero2023VisuoHaptic}. Network $B$ consists of three separate dense layer columns that merge into a dense layer. Positional, force, and temperature values undergo separate processing within these columns.

Both networks were trained with the loss function presented in Equation~\ref{eq_lossRuppel}, which consists of the summation of the mean absolute error (MAE) and the mean squared error (MSE).
\begin{equation} \label{eq_lossRuppel}
    \text{$Loss$}= \frac{\sum_{i=1}^{n}(|y_{i}-\hat{y}_{i}| + (y_{i}-\hat{y}_{i})^{2})}{n}
\end{equation}

The authors report achieving 9.3\% normalized mean absolute error (MAE) over all channels calculated on the z-score normalized outputs of the neural network, i.e., not in the original scale between 0 and 4095, with Network $B$. However, Network $A$ reached 11.3\% MAE. In both cases, the reported results values consider only one of the two dynamic pressure channels (\textit{$pac_0$}) because both channels are nearly identical. The reported results also omitted the heat flow channel (\textit{tac}). During the neural network training phase, both channels, i.e., \textit{$pac_1$} and \textit{tac}, were utilized without justification for this approach, resulting in an output vector comprising 23 channels.

Following their findings, Network $B$ with separate columns yielded better results and had less trainable parameters. Thus, we use it in all our subsequent investigations. In particular, we identify two possible improvements: one related to the temperature used as an input, since this information is unavailable in simulators. The second area of improvement concerns the windowing size used for the force and position values used in the original paper, which was not sufficiently justified.

We use a similar split used by Ruppel et al.\cite{Ruppel2018} to analyze these possible improvements. The test and validation sets consist of 30 chunks of data, each comprising 1000 data points. This configuration ensures that each set, independently, accounts for 10\% of the entire dataset. We train the network without early stopping for 50 epochs. We also conduct 10-fold cross-validation to enhance the statistical robustness of our metric reporting. We also do not consider the channels \textit{$pac_1$} and \textit{tac} in the metric calculation.

\subsection{Investigating the Training with Temperature Readings}
\begin{figure}[htbp]
    \centering
    \includegraphics[width=\columnwidth]{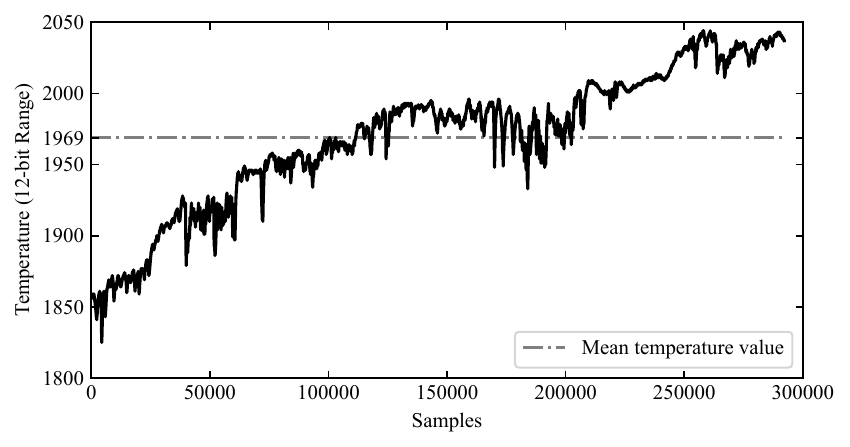}
    \caption{Temperature values of the BioTac sensor in the Ruppel et al.~\cite{Ruppel2018}'s dataset. The dashed black line represents the average temperature value.}
    \label{fig:temperature_values}
\end{figure}

The authors used the current temperature readings (\textit{tdc}) at timestep $t=T$ of the BioTac sensor to predict the output values. Since the used simulation environment Gazebo~\cite{Koenig2004gazebo} lacks temperature information, they suggest fixing that input value to a specific constant, specifically the mean temperature value of the entire collected dataset. 
However, the BioTac sensor is sensitive to temperature changes~\cite{Wettels2007}, and plotting the temperature values of the dataset, Figure~\ref{fig:temperature_values}, it can be seen that the temperature readings of the BioTac sensor continuously increase over time. This data suggests that fixing the temperature to a specific constant value will decrease the model's performance.

\begin{figure}[htbp]
    \centering
    \includegraphics[width=\columnwidth]{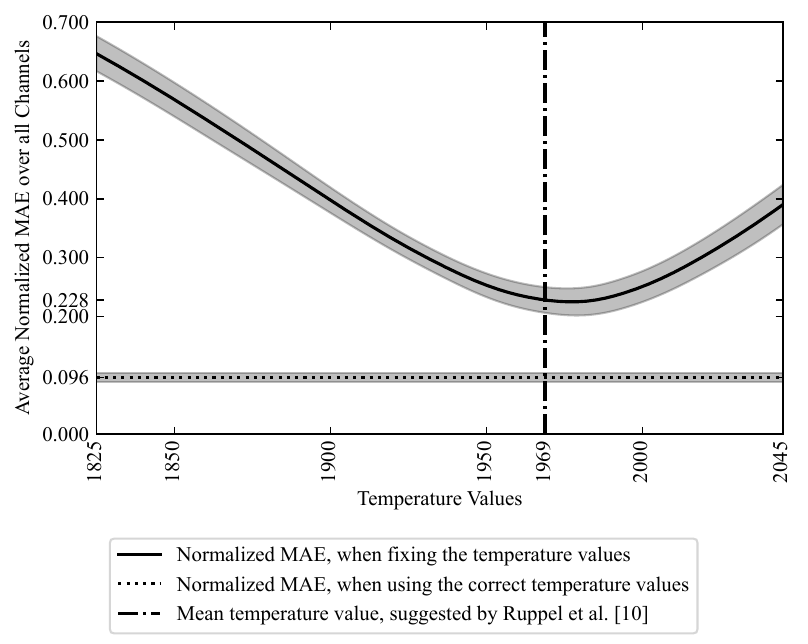}
    \caption{Average normalized MAE over all channels for ten folds for Network $B$. The solid line represents the normalized MAE values when fixing the temperature input value and probing all temperatures in the dataset. The dashed line represents the normalized MAE, when using the appropriate correct temperature values of the test set. The shaded area represents the upper and lower limits calculated by the standard deviation. The vertical dash-dotted line represents the mean temperature of the entire dataset.}
    \label{fig:fix_temp}
\end{figure}

To test the effect of fixing the temperature value, we train Network $B$ using the same hyperparameters used by the authors. However, during the testing phase, we first compute the network's outputs using the correct temperature values provided in the dataset. Second, we compute the network's output using a fixed temperature value. Here, we test all temperature values in the dataset, including the mean value suggested by Ruppel et al.~\cite{Ruppel2018}. The results are shown in Figure~\ref{fig:fix_temp}.

Examining Figure~\ref{fig:fix_temp}, we observe the lowest error when using the correct temperature as input, $9.6\%$. However, fixing the temperature in the best-case scenario leads to an averaged normalized MAE of $22.8\%$. Hence, fixing the input temperature value translates to a relative performance loss of $24.4\%$. We determine the relative loss between the two scenarios by determining the MAE of a naive model ($53.7\%$), which always outputs the mean value of each channel as the prediction.
These findings emphasize the need for an improved solution focusing on learning the data outputs without relying on temperature information since these are unavailable in the simulation environment.

\subsection{Investigating the Windowing Size}\label{sec:investigation_window_size}
The authors reported using the input vector force readings at three timesteps: $T-10$, $T$, and $T+10$, omitting all intermediate timesteps and claiming their inclusion would lead to overfitting. Moreover, the use of force values from future timesteps, i.e., $T+10$, remains unexplained, and error reduction by including future values was not quantified. Thus, we conducted a thorough analysis of the choice of windowing size for force values and explored the impact of incorporating force values from future timesteps. 
To investigate this, we re-train Network $B$ across six different conditions, keeping the current temperature and current position and varying in each experiment the windowing size of the force values as follows:
\begin{enumerate}
    \item Forces $F_t$: $ \forall \text{ } t \in\{T, T-10, T+10\}$.
    \item Forces $F_t$: $ \forall \; t \in\{T, T-10\}$.
    \item Forces $F_t$: $ \forall \; t \in\{T\}$.
    \item Forces $F_t$: $ \forall \text{ } t \in\{T, T-10, T-5, T+5, T+10\}$.
    \item Forces $F_t$: $ \forall \text{ } t \in[T-10, T+10], \quad t \in \mathbb{N}$.
    \item Forces $F_t$: $ \forall \text{ } t \in[T-10, T], \quad t \in \mathbb{N}$.
\end{enumerate}

\begin{figure}[htbp]
    \centering
    \includegraphics[width=0.95\columnwidth]{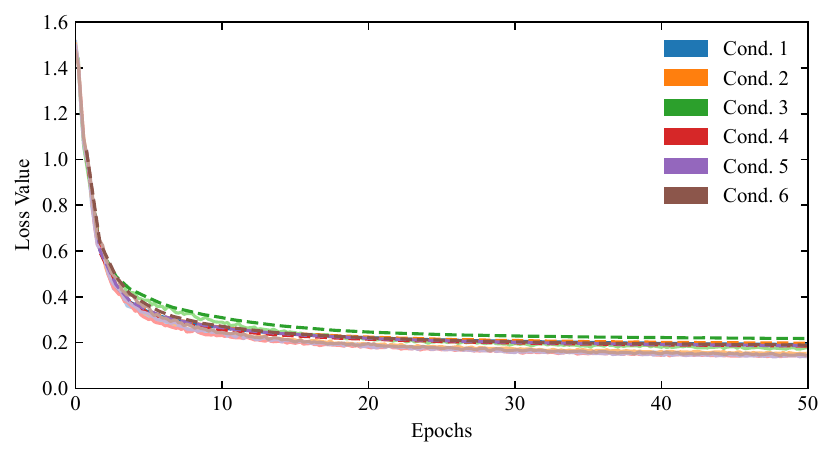}
    \caption{Visualization of the training and validation loss values for all investigated conditions, showing no sign of overfitting. The solid line represents the mean training loss function over all ten folds. The dashed lines depict the mean validation loss over all ten folds.}
    \label{fig:trainValLoss}
\end{figure}

The training and validation loss curves for all conditions are depicted in Figure~\ref{fig:trainValLoss}. Additionally, Table~\ref{tab:WindowInvestigation} illustrates these six conditions' normalized mean absolute errors.

\begin{table}[ht]
\renewcommand{\arraystretch}{1.20}
\centering
\caption{Results of all trained models over all ten folds. The value in parentheses represents the standard deviation. MAE is calculated on the output values in the original scale. Norm. MAE is calculated on the normalized output of the network; the lower, the better.}\label{tab:WindowInvestigation} 
\setlength\tabcolsep{4pt} 
\begin{tabular}{|l|c|c|c|c|c|} 
\hline
\multicolumn{1}{|l|}{\multirow{2}{*}{}} &  \multicolumn{1}{c|}{\multirow{2}{*}{Nb.\ Param.}} & MAE & \multicolumn{1}{c|}{\begin{tabular}[c]{@{}c@{}}Norm.\\ MAE\end{tabular}}     & MAE & \begin{tabular}[c]{@{}c@{}}Norm. \\ MAE\end{tabular}    \\ \cline{3-6} 
\multicolumn{1}{|l|}{}& \multicolumn{1}{|c|}{}& \multicolumn{2}{c|}{Over all Channels} & \multicolumn{2}{c|}{Electrodes Only}\\ \hline
\multicolumn{1}{|l|}{\multirow{2}{*}{1}} & \multicolumn{1}{c|}{\multirow{2}{*}{806K}} & 18.977 & 0.228 & 17.147 & 0.232 \\
& &(1.719) & (0.022) & (1.642) & (0.023) \\\hline
\multicolumn{1}{|l|}{\multirow{2}{*}{2}} & \multicolumn{1}{c|}{\multirow{2}{*}{805K}} & 23.220 & 0.237 & 18.941 & 0.239 \\
& &(1.403) & (0.018) & (1.458) & (0.020) \\\hline
\multicolumn{1}{|l|}{\multirow{2}{*}{3}} & \multicolumn{1}{c|}{\multirow{2}{*}{804K}} & 25.739 & 0.245 & 19.012 & 0.240 \\
& &(1.539) & (0.019) & (1.498) & (0.021) \\\hline
\multicolumn{1}{|l|}{\multirow{2}{*}{4}} & \multicolumn{1}{c|}{\multirow{2}{*}{807K}} & 21.944 & 0.233 & 19.003 & 0.240 \\
& &(1.514) & (0.019) & (1.473) & (0.020) \\\hline
\multicolumn{1}{|l|}{\multirow{2}{*}{5}} & \multicolumn{1}{c|}{\multirow{2}{*}{819K}} & 21.809 & 0.234 & 19.194 & 0.242 \\
& &(1.548) & (0.020) & (1.529) & (0.021) \\\hline
\multicolumn{1}{|l|}{\multirow{2}{*}{6}} & \multicolumn{1}{c|}{\multirow{2}{*}{812K}} & 22.319 & 0.234 & 19.038 & 0.240 \\
& &(1.469) & (0.019) & (1.468) & (0.021) \\\hline
\end{tabular}
\end{table}

Figure~\ref{fig:trainValLoss} shows no sign of overfitting when using shorter intervals for the force values, as both training and validation loss curves converge over time. Table~\ref{tab:WindowInvestigation} reveals no apparent decrease in the metrics between all six input vector combinations when testing on a fixed temperature value equal to the mean temperature of the entire dataset. However, relying solely on the mean MAE over ten folds for comparison may be misleading. A more accurate approach is to compute paired differences of normalized MAE for each fold separately and average these differences. This ensures a fair comparison, as the same ten folds are consistently used across all combination experiments. 

We calculate the paired differences for the input combination used by the authors against all other investigated combinations. Statistical assessment is conducted using a left-tailed paired $t$-test. Since cross-validation with a single dataset involves reusing different training data points in multiple folds, it violates the independence assumptions of the paired $t$-test~\cite{Dietterich1998}. Hence, we employ the corrected paired $t$-test by Nadeau and Bengio~\cite{Nadeau_2003_inference}, which accounts for the number of training and test data points and addresses the issue of high type I errors in the common paired $t$-test. The null hypothesis assumes that the paired difference is equal to $0$, indicating no differences. The alternative hypothesis suggests that the normalized MAE of the first distribution is lower than that of the second distribution. We report the paired normalized MAE difference and the $p$-value in Table~\ref{tab:t_test_ruppel}.

\begin{table}[htb]
\renewcommand{\arraystretch}{1.20}
\centering
\caption{Significance Test with the corrected paired $t$-test~\cite{Nadeau_2003_inference} conducted on different input combinations pairs. The first value depicts the paired normalized MAE difference in percent over the ten folds, and the second value between parenthesis represents the $p$-value.}
\label{tab:t_test_ruppel}
\setlength\tabcolsep{4pt}
\begin{tabular}{|l|c|c|c|c|c|}
\hline
 & 1 vs 2 & 1 vs 3 & 1 vs 4 & 1 vs 5 & 1 vs 6 \\ \hline
Paired MAE Diff.  & -0.893\% & -1.763\% & -0.513\% & -0.630\% & -0.655\% \\
$p$-value     & \textbf{(0.004)} & \textbf{(0.000)} & \textbf{(0.044)} & \textbf{(0.009)}& \textbf{(0.013)}  \\ \hline
\end{tabular}
\end{table}

Table~\ref{tab:t_test_ruppel} indicates that all comparisons are significant with $p<0.05$.
These results show that adding force values from past and future timesteps, 1 vs 2 and 1 vs 3, decreases the error by ca. $0.9\%$ and $1.8\%$, respectively. In addition, using shorter intervals of $10\;ms$, 1 vs 5, or including force values at timesteps $T-5$ and $T+5$, 1 vs 4, does not reduce the error values. We will still consider these investigations and analyses and check different input combinations for our proposed solutions.

\section{Methodology}
\label{sec:Methodology}
Based on the results presented in the previous section, we identify two areas of improvement for the BioTac simulation. The first concerns using or omitting temperature as an input value. The second concerns the windowing size used for the force and position values. This section describes the methodology used to optimize the BioTac simulation.

\subsection{Training without Temperature Readings}
\label{sec:Methodology_temp}
\begin{figure}[htbp]
    \centering
    \includegraphics[width=0.95\columnwidth]{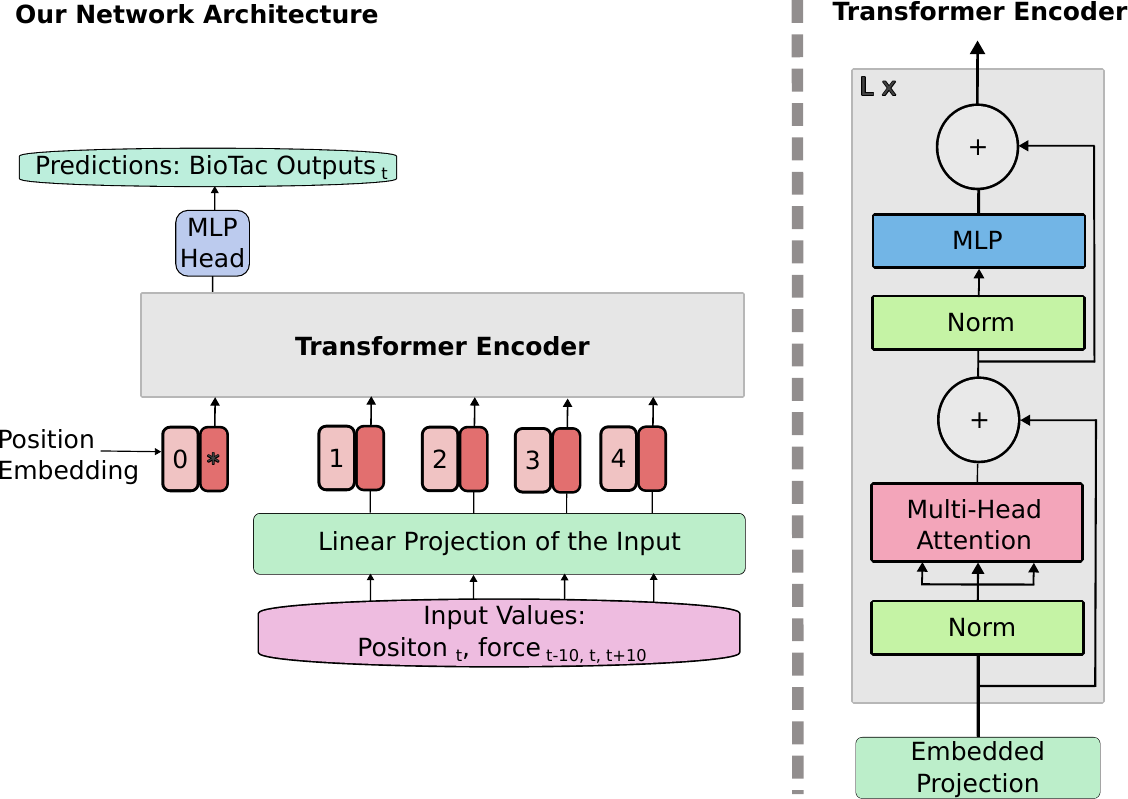}
    \caption{Visualization of our used transformer architecture, based on Dosovitskiy et al.~\cite{Dosovitskiy2021ViT}. L stands for the number of attention blocks used in the transformer encoder.}
    \label{fig:transformer}
\end{figure}

Firstly, we train the baseline network $B$ provided by Ruppel et al.\cite{Ruppel2018} and use the mean temperature value of the dataset as input, and we use these scores as baseline. Next, we implement three approaches: a classical method using XGBoost~\cite{tianqi2016xgboost}, a gradient-boosting algorithm. XGBoost is fast and suitable for predicting continuous output variables~\cite{Bentjac2019ACA}. A feed-forward deep neural network, given that this type of network is also commonly used in regression tasks~\cite{das2020DLRegression}. Finally, we implement a transformer network since these proved beneficial in several time-series tasks~\cite{wen2023transformers}. Inspired by vision transformers (ViT)~\cite{Dosovitskiy2021ViT}, we reformulate the classification task used in the ViT into a regression task by using a transformer encoder to predict the output vector. We present in Figure~\ref{fig:transformer} the used backbone architecture for the transformer network~\cite{Dosovitskiy2021ViT}.

We use Ruppel et al.~\cite{Ruppel2018}'s normalization, input, and output vectors with minor changes. For instance, we use the same input vector but without the temperature value as input resulting in an input vector of 12 values, i.e., $x$, $y$, and $z$ at timestep $T$, force values ($F_{xt}$, $F_{yt}$, and $F_{zt}$) at timesteps $T-10$, $T$, and $T+10$. The output value used consists of all 19 electrodes signals {$e_{1}$…$e_{19}$}, the pressure values $pdc$, and the dynamic vibration $pac_0$ at timestep $T$, i.e., 21 output values. All channels' input and output values are normalized using the z-score normalization.

In contrast to Ruppel et al., ~\cite{Ruppel2018}, we thoroughly fine-tune the hyperparameters of the tested models, namely XGBoost, feed-forward deep neural network, and transformer, using SMAC3~\cite{Lindauer2022SMAC}. The normalized mean absolute error (MAE) is used as a cost function. We randomly split the data into training and validation sets (90\% / 10\%) with random seed values. To select the best configuration, we utilize hyperband~\cite{Lisha2017Hyperband} as an intensifier for the feed-forward network and the transformer, and we use a maximum budget of 30 epochs.
The hyperparameters and their corresponding search space are reported in the supplementary material.

XGBoost does not support multi-variable output vectors. Thus, we fit one regressor for each channel. All regressors use the same hyperparameters determined with SMAC3. During training, we utilize the MAE loss function.
The feed-forward network and the transformer are optimized using Adam~\cite{Kingma2015Adam}, and we use the loss function presented in Equation~\ref{eq_lossRuppel}.

Once the hyperparameters are determined, we use them to train our models from scratch. We split the dataset into training, validation, and test set in 80\% / 10\% / 10\%, respectively. We perform early stopping on the validation loss to avoid overfitting. In contrast, k-fold cross-validation with ten folds is executed for a robust test set evaluation.

\subsection{Varying the Windowing Size}
We also thoroughly investigate the choice of the input window for all tested approaches while omitting the temperature values. Beyond adjusting the force values sampling interval proposed in Section~\ref{sec:investigation_window_size}, we explore two additional input combinations. For both input combinations 1 and 5, we include previous and next positional values within the sampling window to provide the network with contextual touch location information. These are respectively denoted as input combinations 7 and 8:

\begin{enumerate}
    \setcounter{enumi}{6}
    \item Current, last, and next position and forces $F_t$: $ \forall \text{ } t \in\{T, T-10, T+10\}$.
    \item Current, last, and next position and forces $F_t$: $ \forall \text{ } t \in[T-10, T+10], \quad t \in \mathbb{N}$.
\end{enumerate}

We only test within the timestep range of $t-10$ to $t+10$ since the force values' sampling frequency of 100 Hz is a multiple of 10, which aligns with the position's sampling frequency of 10 Hz. 
%
%
We use the SMAC procedure described in the previous Subsection~\ref{sec:Methodology_temp} with the same hyperparameter search space. More details are in the supplementary material.

\section{Results}
Here, we first report the results of our tested approaches that are trained without temperature. Next, we delve into a detailed investigation of a better input window.

\subsection{Training without Temperature Readings}
After executing SMAC for all approaches, the models are trained, and the mean absolute error is calculated. 
We also determined the number of learnable parameters. For XGBoost regressors, we calculate the number of parameters by counting the number of nodes in all regressors across channels after training them. We then average this count over the folds, accounting for variability across each fold. The results are outlined in Table~\ref{tab:table_Results_1}.

\begin{table}[ht]
\renewcommand{\arraystretch}{1.20}
\centering
\caption{Results of the all trained models over all ten folds. The value in parentheses represents the standard deviation. The metrics are calculated over all channels and over all electrodes. MAE is calculated on the output values in the original scale. Norm. MAE is calculated on the normalized output of the network; the lower, the better.}
\label{tab:table_Results_1}
\setlength\tabcolsep{4pt} 
\begin{tabular}{|l|c|c|c|c|c|}
\hline
& Nb. &\multicolumn{1}{|c|}{\multirow{2}{*}{MAE}} & Norm. & \multicolumn{1}{|c|}{\multirow{2}{*}{MAE}} & Norm. \\
& Param. & & MAE & & MAE \\ \cline{3-6} 
& &\multicolumn{2}{|c|}{Over all Channels} & \multicolumn{2}{|c|}{Electrodes Only} \\ \hline

\multicolumn{1}{|l|}{\multirow{2}{*}{Ruppel et al.~\cite{Ruppel2018}}} &  \multicolumn{1}{|c|}{\multirow{2}{*}{806K}}  & 18.977 & 0.228 & 17.147 & 0.232 \\
& & (1.719) & (0.022) & (1.642) & (0.023) \\ \hline

Our XGBoost &  \multicolumn{1}{|c|}{\multirow{2}{*}{1584K}}  & \textbf{13.368} & \textbf{0.150} & \textbf{11.446} & \textbf{0.150} \\
Regressor & & (1.340) & (0.014) & (1.204) & (0.015)   \\ \hline

Our Feed-Forward &  \multicolumn{1}{|c|}{\multirow{2}{*}{2233K}}  & 14.693 & 0.168 & 12.724 & 0.169 \\
Neural Network & & (1.386) & (0.015) & (1.252) & (0.016)  \\ \hline

Our Transformer &  \multicolumn{1}{|c|}{\multirow{2}{*}{\textbf{599K}}}  & 13.760 & 0.156 & 11.736 & 0.155 \\
Encoder & & (1.400) & (0.016) & (1.280) & (0.016) \\ \hline
\end{tabular}
\end{table}

From Table~\ref{tab:table_Results_1}, it can be seen that the XGBoost and transformer approaches without temperature readings achieved a statistically significant improvement compared to the baseline. The absolute improvement measured in the normalized MAE ranges from $6.0\%$ to $7.8\%$. The transformer encoder stands out for its compactness and reduced number of trainable parameters. While the XGBoost reaches the lowest normalized mean absolute error with $15.0\%$.

\begin{figure}[tbp]
    \centering
    \includegraphics[width=\columnwidth]{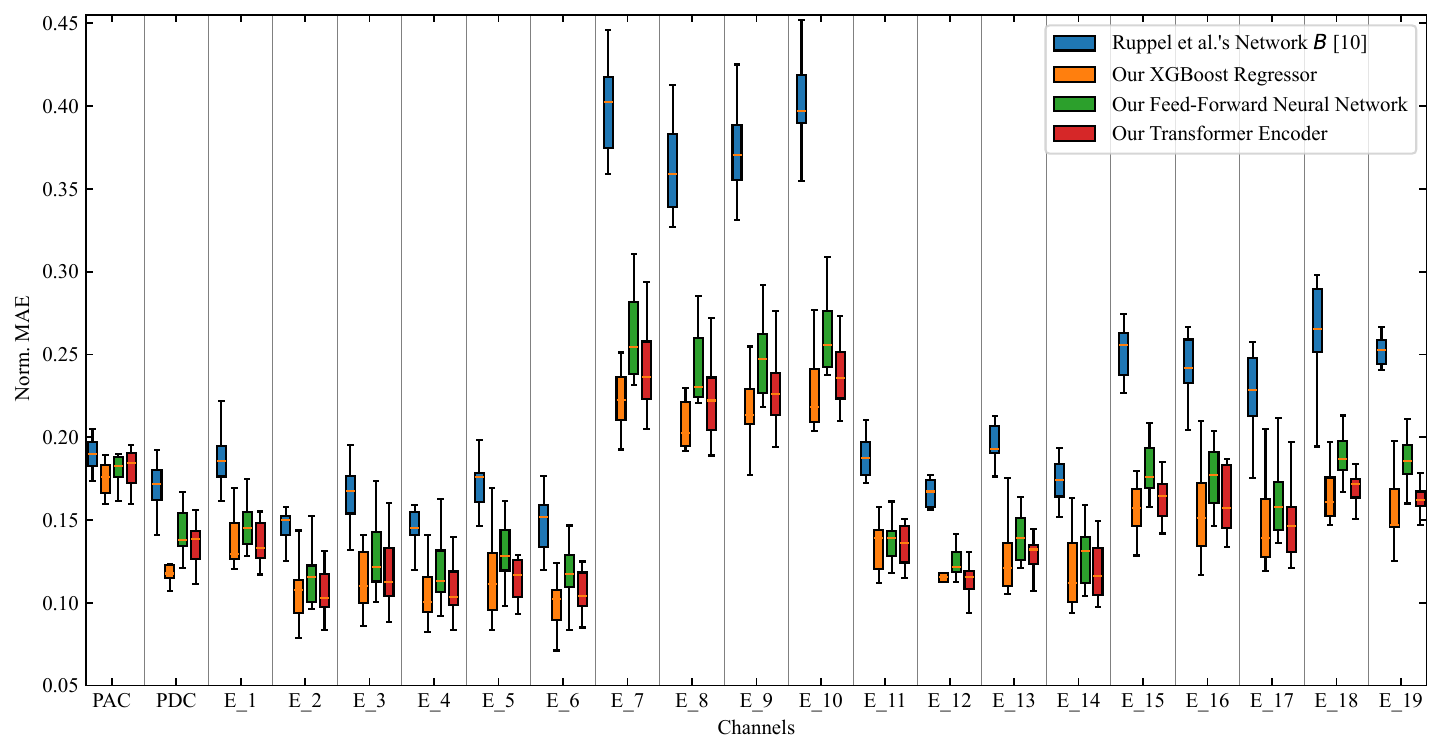}
    \caption{Distribution of the normalized MAE for each channel and all models.}
    \label{fig:boxplots_temp}
\end{figure}

We also present the normalized MAE for each channel in Figure~\ref{fig:boxplots_temp}. There, it can be seen that the prediction errors differ substantially between channels. Particularly, electrodes 7, 8, 9, and 10, located at the tip of the BioTac sensor, result in the highest errors. This phenomenon is also seen for the XGBoost regressors, which have a dedicated regressor for each channel. This suggests a high non-linear dynamic of those electrodes.

\begin{figure}[tbp]
    \centering
    \includegraphics[width=\columnwidth]{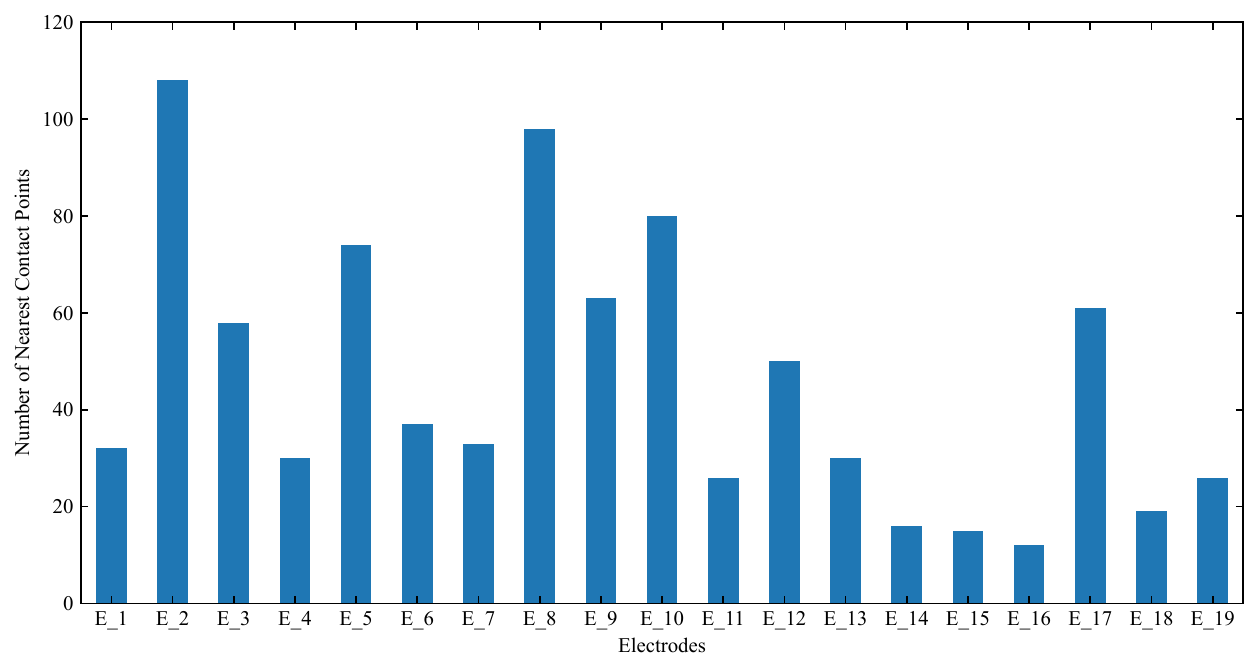}
    \caption{Distribution of nearest contact points for each electrode.}
    \label{fig:contact_hist}
\end{figure}

Another possible reason for the differences in prediction errors could be the different number of samples available for each channel; i.e., the dataset might be unbalanced. We assessed that by selecting all data points at the beginning of a contact cycle having a force measurement higher than $0.3\; N$, located close to the BioTac sensor surface within a distance of less than $2\;mm$. Subsequently, we identify the nearest electrode to each contact point using the 3D spatial positions of each electrode provided by Lin et al.~\cite{Lin2013}. The results are depicted in Figure~\ref{fig:contact_hist}.

Figure~\ref{fig:contact_hist} highlights a non-uniform distribution of the number of touches near each electrode, implying an unbalanced dataset. However, this may not be the only reason contributing to the high MAE for specific electrodes.
Based on the work of Lin et al.~\cite{Lin2013}, we know that the BioTac sensor does not have radial symmetry. In addition, the fluid volume is not the same throughout the sensor. Particularly near electrodes 7, 8, 9, and 10, there is a higher fluid volume, which is partially visible in Figure~\ref{fig:biotacFrame_b}. This supports the hypothesis that some of the error is due to the non-linear dynamics of the sensor.

\subsection{Varying the Windowing Size}
After executing SMAC for all approaches, the models are trained, and the mean absolute error is calculated. The best hyperparameters for each trained model and input combination and the result metrics values are reported in the supplementary material.

Figure~\ref{fig:boxplots_exp} presents the distribution of the normalized MAE for all combinations across all approaches. 
XGBoost attains the lowest MAE when trained with input combination 8, incorporating force values with shorter intervals and the last and next position, reaching a value of 14.8\%, which represents an $8.0\%$ improvement over the baseline. Likewise, our transformer encoder achieves a $7.8\%$ improvement over the baseline with the same input combination.

\begin{figure}[tbp]
    \centering
    \includegraphics[width=\columnwidth]{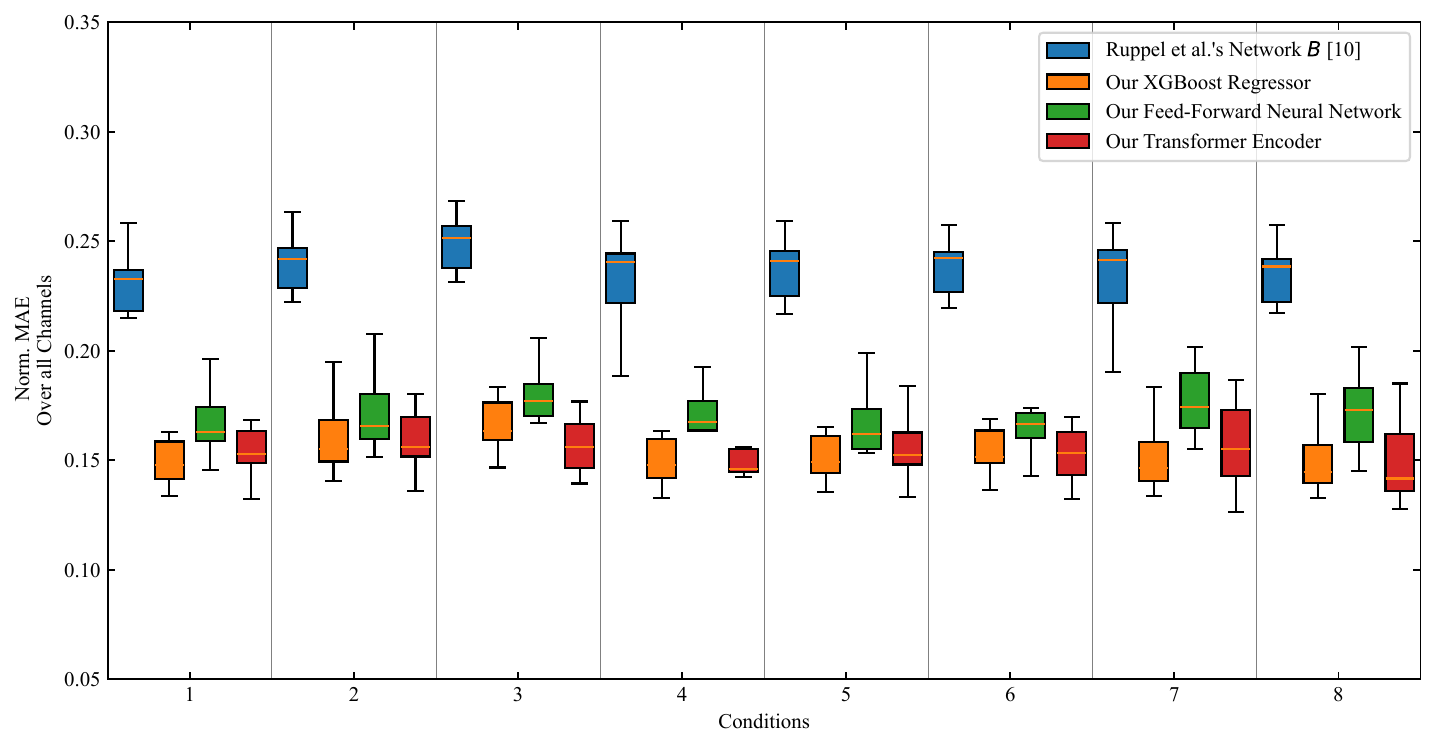}
    \caption{The Normalized MAE was calculated over all channels for the four models and all input combinations.}
    \label{fig:boxplots_exp}
\end{figure} 

We also conduct a corrected paired significance $t$-test~\cite{Nadeau_2003_inference} and report the paired normalized MAE difference and $p$-value for three specific comparison combinations: 5 vs 1, 7 vs 1 and 8 vs 1. The null hypothesis assumes that the paired difference is equal to $0$, indicating no differences. The alternative hypothesis suggests that the normalized MAE of the first distribution is lower than that of the second distribution. Comparison 5 vs 1 assesses the impact of using force values of shorter intervals on error reduction, 7 vs 1 examines the effect of including the last and next different position, and 8 vs 1 evaluates the significance of incorporating more force values at shorter intervals along with the last and next position in error reduction. 
The paired normalized MAE difference, $t$-statistic and $p$-values for various combination pairs across all models are reported in the supplementary material.

Table~\ref{tab:t_test_our_networks} indicates that adding more force values of shorter intervals does not directly lead to a reduction in error (5 vs 1), and the same applies when adding the last and next contact position (7 vs 1). However, combining both information yielded statistically significant improvements for the XGBoost regressor and the transformer encoder in the 8 vs 1 comparison. However, they are small, ranging between 0.2\% and 0.6\%.

\begin{table}[ht]
\renewcommand{\arraystretch}{1.20}
\centering
\caption{Significance Test with the corrected paired $t$-test~\cite{Nadeau_2003_inference} conducted on different input combinations. The first value depicts the paired normalized MAE difference in percent over the ten folds, and the second value between parenthesis represents the $p$-value.}
\label{tab:t_test_our_networks}
\setlength\tabcolsep{4pt}
\begin{tabular}{|l|c|c|c|}
\hline
 & 5 vs 1 & 7 vs 1 & 8 vs 1   \\ \hline
                                                                       
                                                            Our XGBoost    & 0.282\% & -0.055\% & -0.208\% \\
 Regressor   & (0.999) & (0.190) & \textbf{(0.009)} \\\hline

                                                            Our Feed-Forward & 0.217\% & 0.862\%& 0.402\% \\
                                                            Neural Network & (0.715) & (0.961) & (0.758) \\\hline

                                                            Our Transformer  & 0.077\% & 0.188\% & -0.662\% \\
                                                            Encoder  & (0.631) & (0.672) & \textbf{(0.047)} \\\hline
\end{tabular}
\end{table}

\begin{table}[ht]
\renewcommand{\arraystretch}{1.20}
\centering
\caption{Significance Test with the corrected paired $t$-test~\cite{Nadeau_2003_inference} conducted for all models. The first value depicts the paired normalized MAE difference value in percent, and the second value between parenthesis represents the $p$-value.}
\label{tab:t_test_all_networks}
\setlength\tabcolsep{4pt}
\begin{tabular}{|l|c|c|c|}
\hline
\multicolumn{1}{|l|}{\multirow{2}{*}{vs}} & Our XGBoost & Our XGBoost &  Our Transformer   \\ 
& Our FFNN &  Our Transformer &  Our FFNN  \\ \hline
\multicolumn{1}{|l|}{\multirow{2}{*}{1}} 
 & -1.778\%& -0.531\% & -1.247\%\\
 & \textbf{(0.000)} & (0.074) & \textbf{(0.002)} \\\hline
\multicolumn{1}{|l|}{\multirow{2}{*}{2}} 
 & -1.360\% & -0.259\%& -1.101\% \\
 & \textbf{(0.001)} & (0.250) & \textbf{(0.000)} \\\hline
\multicolumn{1}{|l|}{\multirow{2}{*}{3}} 
 & -1.272\% & 0.851\% & -2.123\% \\
 & \textbf{(0.003)} & (0.996) & \textbf{(0.000)} \\\hline
\multicolumn{1}{|l|}{\multirow{2}{*}{4}} 
 & -1.989\%& -0.032\% & -1.956\%\\
 & \textbf{(0.000)} & (0.459) & \textbf{(0.000)} \\\hline
\multicolumn{1}{|l|}{\multirow{2}{*}{5}} 
 & -1.279\%& -0.173\% & -1.107\% \\
 & \textbf{(0.002)} & (0.322) & \textbf{(0.002)} \\\hline
\multicolumn{1}{|l|}{\multirow{2}{*}{6}} 
 & -1.106\% & 0.092\% & -1.197\% \\
 & \textbf{(0.003)} & (0.616) & \textbf{(0.002)} \\\hline
\multicolumn{1}{|l|}{\multirow{2}{*}{7}} 
 & -2.696\%& -0.774\% & -1.921\% \\
 & \textbf{(0.000)} & (0.105) & \textbf{(0.000)} \\\hline
\multicolumn{1}{|l|}{\multirow{2}{*}{8}} 
 & -2.389\%& -0.077\% & -2.312\% \\
 & \textbf{(0.001)} & (0.439) & \textbf{(0.002)} \\\hline

\end{tabular}
\end{table}

We also analyzed the significance between the tested approaches for all input combinations and reported the results. 
Table~\ref{tab:t_test_all_networks} reveals that both the XGBoost regressor and the transformer encoder outperform the feed-forward neural network across all input combinations, demonstrating an error reduction ranging between $1.1\%$ and $2.7\%$. However, a direct comparison between the XGBoost and transformer yields no significant values, suggesting no difference between the paired MAE.

\begin{figure}[htbp]
    \centering
    \includegraphics[width=\columnwidth]{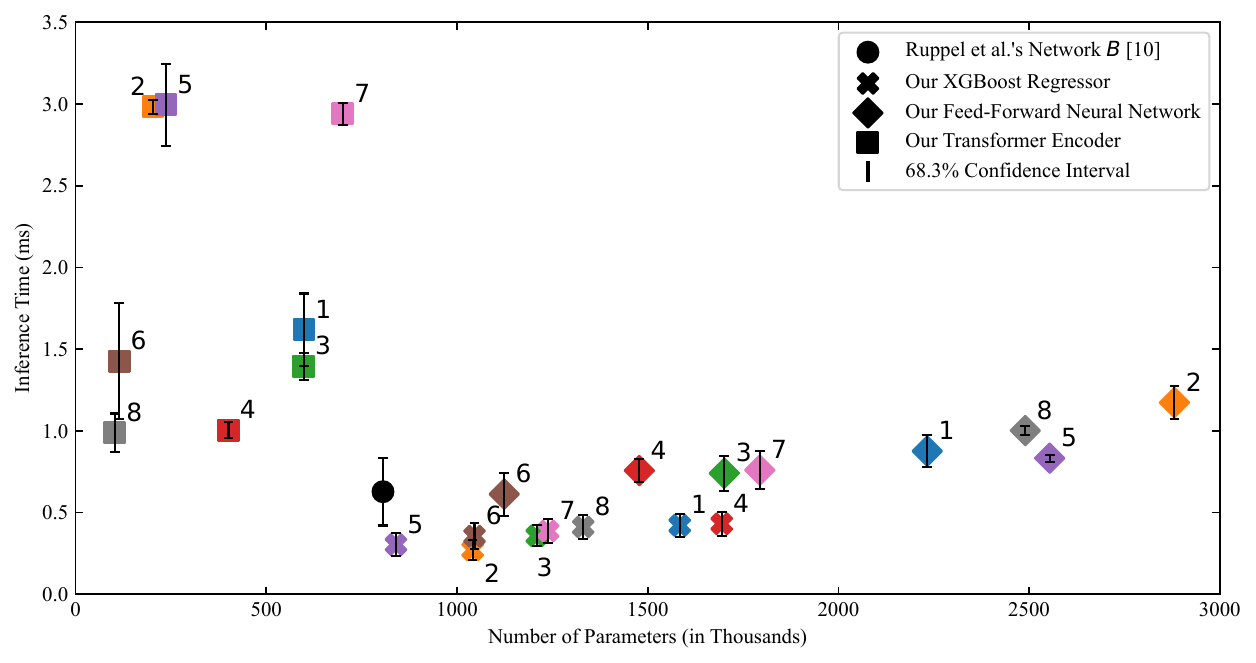}
    \caption{Visualization of inference time against the number of parameters for each model over all input combinations.}
    \label{fig:inf_param}
\end{figure}

Considering the importance of inference time during the deployment phase, we compute the inference time over 100 random inputs for all approaches and subsequently average the results. We run the inference calculation on a system with an \textit{AMD Ryzen 7 pro 4750u} CPU. Figure~\ref{fig:inf_param} shows the inference time against the number of learnable parameters for all approaches and input combinations. For Ruppel et al.~\cite{Ruppel2018} baseline network, we only plot results for the first input combination, as all other combinations do not differ meaningfully in inference time or numbers of parameters.

Figure~\ref{fig:inf_param} reveals that the transformer network has the lowest number of parameters compared to all other approaches, with an inference time ranging from $1.0\;ms$ up to $3.0\;ms$, i.e., at least two to five times slower than the baseline. 
This is attributed to the quadratic time complexity of the attention blocks~\cite{Yi2023Transform}. XGBoost is the fastest of the tested methods, with approximately half the inference time of the baseline network. 

Considering that the number of floating-point operations per second (FLOPS) offers additional valuable insights into the energy consumption and inference efficiency of neural networks ~\cite{Tang2017Flops}, we include the FLOPS count for all the neural networks, i.e., the baseline network, our feed-forward network, and our transformer encoder, in our supplementary material.

\section{Conclusion}
We present optimizations to simulate tactile signals from the BioTac 2P sensor. Our contributions include a thorough analysis of Ruppel et al.'s work~\cite{Ruppel2018} and a thorough analysis of three alternative solutions. We focused on two areas: the use of temperature readings in the training pipeline of the network, and the choice of other variables in the input vector and the windowing. 

We introduced three alternative models, i.e., an XGBoost regressor, a feed-forward neural network, and an adapted transformer encoder. 
Our results demonstrate that XGBoost and the transformer encoder can achieve significantly lower error values than the baseline. The absolute error reduction stands at $8.0\%$, while the relative error shows a statistically significant improvement of $14.9\%$. 
Furthermore, our investigations reveal a statistically significant improvement when incorporating force values from future timesteps and previous and next position values. Albeit, there are small between 0.2\% up to 0.6\% depending on the model used.

Notably, the XGBoost regressor has the lowest inference time, making it preferable for scenarios where simulation inference time is critical. Despite having less trainable parameters than all other models, the transformers exhibit the highest inference time of the tested models. 

We also report on the limitations of the dataset. Particularly, the dataset is unbalanced and only includes a single indenter type. However, some of the errors can be attributed to the non-linear dynamics of the sensor, probably caused by the non-radial symmetry and non-uniform fluid volume throughout the sensor. 

As future work, the dataset needs to be improved and extended to include different BioTac 2P sensors, varied surrounding temperatures, and several indenter shapes to enhance model robustness and generalizability, mitigating the absence of temperature information in the simulation environment and other non-linear dynamics.

Due to the differentiated performance of different electrodes due to non-linearities and unbalanced dataset, we also suggest training an ensemble of transformer networks to better deal with the non-linear dynamic of the sensor.



\bibliographystyle{IEEEtran}
\bibliography{references.bib}


\begin{table*}[ht]
\centering
\caption{Summary of the hyperparameter search space used for the XGBoost Regressor, when running SMAC3~\cite{Lindauer2022SMAC} and the selected configuration for all input combinations.}\label{tab:hyperparameters1}
\setlength\tabcolsep{4pt} 
\begin{tabular}{|l|ccccccccc|}
\hline
\multirow{2}{*}{Hyperparameters} & \multicolumn{9}{c|}{Our XGBoost Regressor}                                                                                                                                                   \\ \cline{2-10} 
                                 & \multicolumn{1}{c|}{Search Space} & \multicolumn{1}{c|}{Input 1} & \multicolumn{1}{c|}{Input 2} & \multicolumn{1}{c|}{Input 3} & \multicolumn{1}{c|}{Input 4} & \multicolumn{1}{c|}{Input 5} & \multicolumn{1}{c|}{Input 6} &\multicolumn{1}{c|}{Input 7} & Input 8 \\ \hline

eta                      & \multicolumn{1}{c|}{\begin{tabular}[c]{@{}c@{}}Uniform Float\\ $\in$ {[}0.0001, 0.5{]} \end{tabular}}                        &    \multicolumn{1}{c|}{0.0431}       & \multicolumn{1}{c|}{0.06818}  &   \multicolumn{1}{c|}{0.04942}   &   \multicolumn{1}{c|}{0.04457}    &     \multicolumn{1}{c|}{0.07178}    &  \multicolumn{1}{c|}{0.06647}      & \multicolumn{1}{c|}{0.04901}        &    0.04798     \\ \hline

gamma                                & \multicolumn{1}{c|}{\begin{tabular}[c]{@{}c@{}}Uniform Integer\\ $\in$ {[}0, 10{]}    \end{tabular}}            &  \multicolumn{1}{c|}{1}    &   \multicolumn{1}{c|}{7}  &  \multicolumn{1}{c|}{2} &   \multicolumn{1}{c|}{1}   &  \multicolumn{1}{c|}{7}  &   \multicolumn{1}{c|}{3} &      \multicolumn{1}{c|}{3}  &     2  \\ \hline

number estimate                                & \multicolumn{1}{c|}{\begin{tabular}[c]{@{}c@{}}Uniform Integer\\ $\in$ {[}100, 1000{]}    \end{tabular}}        &    \multicolumn{1}{c|}{972}     &     \multicolumn{1}{c|}{155}  &  \multicolumn{1}{c|}{932}  &   \multicolumn{1}{c|}{880} &   \multicolumn{1}{c|}{230}    &     \multicolumn{1}{c|}{913}    &  \multicolumn{1}{c|}{715}  &    949    \\ \hline

max depth                                & \multicolumn{1}{c|}{\begin{tabular}[c]{@{}c@{}}Uniform Integer\\ $\in$ {[}1, 10{]}    \end{tabular}}   &   \multicolumn{1}{c|}{10}     &   \multicolumn{1}{c|}{10} &  \multicolumn{1}{c|}{10}    &   \multicolumn{1}{c|}{10}  &   \multicolumn{1}{c|}{10} &     \multicolumn{1}{c|}{10}    &  \multicolumn{1}{c|}{10}     &     10   \\ \hline

min child weight                                & \multicolumn{1}{c|}{\begin{tabular}[c]{@{}c@{}}Uniform Integer\\ $\in$ {[}1, 100{]}    \end{tabular}} &   \multicolumn{1}{c|}{95} &     \multicolumn{1}{c|}{6} &   \multicolumn{1}{c|}{87}  &   \multicolumn{1}{c|}{67} &    \multicolumn{1}{c|}{59}   &    \multicolumn{1}{c|}{83}     &  \multicolumn{1}{c|}{84}     &    84    \\ \hline

max delta step                                & \multicolumn{1}{c|}{\begin{tabular}[c]{@{}c@{}}Uniform Integer\\ $\in$ {[}0, 10{]}    \end{tabular}}    &     \multicolumn{1}{c|}{7}    &     \multicolumn{1}{c|}{10}   &     \multicolumn{1}{c|}{0} &   \multicolumn{1}{c|}{1}  &   \multicolumn{1}{c|}{5}    &     \multicolumn{1}{c|}{3}    &  \multicolumn{1}{c|}{10}     &   1     \\ \hline

subsample                               & \multicolumn{1}{c|}{\begin{tabular}[c]{@{}c@{}}Uniform Float\\ $\in$ {[}0.5, 1{]}    \end{tabular}}         &     \multicolumn{1}{c|}{0.647}    &   \multicolumn{1}{c|}{0.9632} &   \multicolumn{1}{c|}{0.5042}  &   \multicolumn{1}{c|}{0.5068}  &    \multicolumn{1}{c|}{0.6114}   &    \multicolumn{1}{c|}{0.5486}     &  \multicolumn{1}{c|}{0.507}   &   0.5038    \\ \hline

colsample bytree                               & \multicolumn{1}{c|}{\begin{tabular}[c]{@{}c@{}}Uniform Float\\ $\in$ {[}0.5, 1{]}    \end{tabular}}      &     \multicolumn{1}{c|}{0.9825}    &   \multicolumn{1}{c|}{0.9517} &   \multicolumn{1}{c|}{0.8671}  &   \multicolumn{1}{c|}{0.868}   &   \multicolumn{1}{c|}{0.7449}    &   \multicolumn{1}{c|}{0.8242}      &  \multicolumn{1}{c|}{0.9927}  &     0.7814   \\ \hline

colsample bylevel                             & \multicolumn{1}{c|}{\begin{tabular}[c]{@{}c@{}}Uniform Float\\ $\in$ {[}0.5, 1{]}    \end{tabular}}       &    \multicolumn{1}{c|}{0.9819}   &   \multicolumn{1}{c|}{0.9223} &    \multicolumn{1}{c|}{0.8572} &   \multicolumn{1}{c|}{0.839}   &   \multicolumn{1}{c|}{0.9264}    &     \multicolumn{1}{c|}{0.7642}    &  \multicolumn{1}{c|}{0.9479}  &  0.8407     \\ \hline

colsample bynode                                & \multicolumn{1}{c|}{\begin{tabular}[c]{@{}c@{}}Uniform Float\\ $\in$ {[}0.5, 1{]}    \end{tabular}}     &   \multicolumn{1}{c|}{0.8042} &   \multicolumn{1}{c|}{0.9155}  &   \multicolumn{1}{c|}{0.6173} &   \multicolumn{1}{c|}{0.7416}  &   \multicolumn{1}{c|}{0.9802}    &   \multicolumn{1}{c|}{0.924}      &  \multicolumn{1}{c|}{0.8621}  &    0.9989    \\ \hline

\end{tabular}
\end{table*}


\begin{table*}[ht]
\centering
\caption{Table summarizing the hyperparameter search space used for the feed-forward deep neural network, when running SMAC3~\cite{Lindauer2022SMAC} and the selected configuration for all input combinations.}\label{tab:hyperparameters2}
\setlength\tabcolsep{0.1pt} 
\begin{tabular}{|l|ccccccccc|}
\hline
\multirow{2}{*}{Hyperparameters} & \multicolumn{9}{c|}{Our Feed-Forward Neural Network}                                                                                                                                                   \\ \cline{2-10} 
                                 & \multicolumn{1}{c|}{Search Space} & \multicolumn{1}{c|}{Input 1} & \multicolumn{1}{c|}{Input 2} & \multicolumn{1}{c|}{Input 3} & \multicolumn{1}{c|}{Input 4} & \multicolumn{1}{c|}{Input 5} & \multicolumn{1}{c|}{Input 6}& \multicolumn{1}{c|}{Input 7} & Input 8 \\ \hline

Batch Size                       & \multicolumn{1}{c|}{\begin{tabular}[c]{@{}c@{}}Categorical\\ {[}256, 512{]}  \end{tabular}}             & \multicolumn{1}{c|}{256}        & \multicolumn{1}{c|}{512}    & \multicolumn{1}{c|}{256}       & \multicolumn{1}{c|}{512}    & \multicolumn{1}{c|}{512}        & \multicolumn{1}{c|}{512}        & \multicolumn{1}{c|}{512}        &     512    \\ \hline

Learning Rate             & \multicolumn{1}{c|}{\begin{tabular}[c]{@{}c@{}}Categorical\\ $a \times e^{-c}$\\ for $a \in \mathbb{N}^+$ and $ \in {[}1,9{]}$ \\ $c \in \mathbb{N}^+$ and $ \in {[}2,5{]}$ \end{tabular} }             & \multicolumn{1}{c|}{0.0003}        & \multicolumn{1}{c|}{0.0002}   & \multicolumn{1}{c|}{0.0005}    & \multicolumn{1}{c|}{0.0006}    & \multicolumn{1}{c|}{0.0003}        & \multicolumn{1}{c|}{0.0006}        & \multicolumn{1}{c|}{0.0004}        &   0.0005      \\ \hline

\begin{tabular}[c]{@{}l@{}}Number\\ of Layers ($L$)\end{tabular}          & \multicolumn{1}{c|}{\begin{tabular}[c]{@{}c@{}}Uniform Int\\ Lower: 4\\ Upper: 12 \end{tabular}}             & \multicolumn{1}{c|}{7}        & \multicolumn{1}{c|}{10}   & \multicolumn{1}{c|}{7} & \multicolumn{1}{c|}{7}     & \multicolumn{1}{c|}{7}        & \multicolumn{1}{c|}{9}        & \multicolumn{1}{c|}{7}        &    8     \\ \hline

\begin{tabular}[c]{@{}l@{}}Numebr of Neurons\\ in Layer i\\ for $i \in {[}0, L{]}$ \end{tabular}    & \multicolumn{1}{c|}{\begin{tabular}[c]{@{}c@{}}Uniform Int \\ Lower: 50\\ Upper: 1000\\ Step: 10 \end{tabular}     }            & \multicolumn{1}{c|}{\begin{tabular}[c]{@{}c@{}}[860, 670, \\160, 580,\\ 900, 1000,\\ 440]\end{tabular}}        & \multicolumn{1}{c|}{\begin{tabular}[c]{@{}c@{}}[590,300,\\820,520,\\90,670,\\850,120,\\330,570]\end{tabular}}    & \multicolumn{1}{c|}{\begin{tabular}[c]{@{}c@{}}[620, 470, \\120, 620,\\ 830, 890,\\ 350]\end{tabular}}       & \multicolumn{1}{c|}{\begin{tabular}[c]{@{}c@{}}[730, 390, \\120, 630,\\ 770, 720,\\ 380]\end{tabular}}     & \multicolumn{1}{c|}{\begin{tabular}[c]{@{}c@{}}[820,740,\\190,740,\\1000,850,\\430]\end{tabular}}        & \multicolumn{1}{c|}{\begin{tabular}[c]{@{}c@{}}[580,710,\\600,170,\\270,350,\\70,780,\\690]\end{tabular}}        & \multicolumn{1}{c|}{\begin{tabular}[c]{@{}c@{}}[580,580,\\160,450,\\920,920,\\380]\end{tabular}}        &  \begin{tabular}[c]{@{}c@{}}[470,600,\\440,830,\\790,900,\\190,270]\end{tabular}       \\ \hline

\begin{tabular}[c]{@{}l@{}}Activation function \\ in Layer i\\ for $i \in {[}0, L{]}$  \end{tabular}  & \multicolumn{1}{c|}{\begin{tabular}[c]{@{}c@{}}Categorical\\ {[}sigmoid,  relu, \\ hardtanh, tanh, \\leakyrelu, elu{]}\end{tabular}}             & \multicolumn{1}{c|}{ \begin{tabular}[c]{@{}c@{}} {[}hardtanh,  tanh,\\ elu,  relu, \\leakyrelu, relu,\\ leakyrelu{]}\end{tabular} }        & \multicolumn{1}{c|}{ \begin{tabular}[c]{@{}c@{}} [elu,
relu,\\
hardtanh,
elu,\\
leakyrelu,
relu,\\
elu,
elu,\\
elu,
leakyrelu]\end{tabular}}   

& \multicolumn{1}{c|}{ \begin{tabular}[c]{@{}c@{}} [tanh,
elu,\\
elu,\\
leakyrelu,\\
leakyrelu,\\
elu,\\
leakyrelu]\end{tabular}}   

& \multicolumn{1}{c|}{ \begin{tabular}[c]{@{}c@{}} [tanh,
tanh,\\
elu,\\
hardtanh,\\
sigmoid,\\
leakyrelu,\\
leakyrelu]\end{tabular}}

& \multicolumn{1}{c|}{\begin{tabular}[c]{@{}c@{}} [tanh,
relu,\\
elu,
leakyrelu,\\
elu,
relu,\\
leakyrelu]\end{tabular}}        & \multicolumn{1}{c|}{\begin{tabular}[c]{@{}c@{}} [tanh,
relu,\\
leakyrelu,
elu,\\
leakyrelu,\\
leakyrelu,\\
leakyrelu,\\
hardtanh,\\
leakyrelu]\end{tabular}}        & \multicolumn{1}{c|}{\begin{tabular}[c]{@{}c@{}} [tanh,
relu,\\
leakyrelu,\\
hardtanh,\\
sigmoid,
relu,\\
leakyrelu]\end{tabular}}        &    \begin{tabular}[c]{@{}c@{}} [tanh,
relu,\\
elu,
leakyrelu,\\
hardtanh,\\
leakyrelu,\\
elu,
leakyrelu]\end{tabular}     \\ \hline

\begin{tabular}[c]{@{}l@{}}Negative Slope\\ for Leakyrelu \end{tabular}                 & \multicolumn{1}{c|}{\begin{tabular}[c]{@{}c@{}}Categorical\\  $a \times e^{-1}$\\ for $a \in \mathbb{N}^+$ and $ \in {[}1,9{]}$ \end{tabular}}             & \multicolumn{1}{c|}{0.5}        & \multicolumn{1}{c|}{0.7}    & \multicolumn{1}{c|}{0.4}    & \multicolumn{1}{c|}{0.2}    & \multicolumn{1}{c|}{0.1}        & \multicolumn{1}{c|}{0.4}        & \multicolumn{1}{c|}{0.3}        &    0.6     \\ \hline

\end{tabular}
\end{table*}


\begin{table*}[ht]
\centering
\caption{Table summarizing the hyperparameter search space used for the transformer encoder, when running SMAC3~\cite{Lindauer2022SMAC} and the selected configuration for all input combinations.}\label{tab:hyperparameters3}
\setlength\tabcolsep{4pt} 
\begin{tabular}{|l|ccccccccc|}
\hline
\multirow{2}{*}{Hyperparameters} & \multicolumn{9}{c|}{Our Transformer Encoder}                                                                                                                                                   \\ \cline{2-10} 
 & \multicolumn{1}{c|}{Search Space} & \multicolumn{1}{c|}{Input 1} & \multicolumn{1}{c|}{Input 2} & \multicolumn{1}{c|}{Input 3} & \multicolumn{1}{c|}{Input 4} & \multicolumn{1}{c|}{Input 5} & \multicolumn{1}{c|}{Input 6} & \multicolumn{1}{c|}{Input 7} & Input 8 \\ \hline

Batch Size                       & \multicolumn{1}{c|}{\begin{tabular}[c]{@{}c@{}}Categorical\\ {[}256, 512{]} \end{tabular}}             & \multicolumn{1}{c|}{512}     & \multicolumn{1}{c|}{512}      & \multicolumn{1}{c|}{512}         & \multicolumn{1}{c|}{256}     & \multicolumn{1}{c|}{512}        & \multicolumn{1}{c|}{256}        & \multicolumn{1}{c|}{512}        &      256   \\ \hline

Learning Rate                                 & \multicolumn{1}{c|}{\begin{tabular}[c]{@{}c@{}}Categorical\\ $a \times e^{-c}$\\ for $a \in \mathbb{N}^+$ and $ \in {[}1,9{]}$ \\  $c \in \mathbb{N}^+$ and $ \in {[}2,5{]}$   \end{tabular}}             & \multicolumn{1}{c|}{0.00009}        & \multicolumn{1}{c|}{0.00004}  & \multicolumn{1}{c|}{0.00009}     & \multicolumn{1}{c|}{0.0004}    & \multicolumn{1}{c|}{0.004}        & \multicolumn{1}{c|}{0.0003}        & \multicolumn{1}{c|}{0.00004}        &   0.002      \\ \hline

\begin{tabular}[c]{@{}l@{}}Number\\ of Layers ($L$)\end{tabular}  & \multicolumn{1}{c|}{  \begin{tabular}[c]{@{}c@{}}Uniform Int\\ Lower: 2\\ Upper: 8\end{tabular}    }             & \multicolumn{1}{c|}{3}        & \multicolumn{1}{c|}{8}      & \multicolumn{1}{c|}{3}      & \multicolumn{1}{c|}{2}    & \multicolumn{1}{c|}{7}        & \multicolumn{1}{c|}{4}        & \multicolumn{1}{c|}{7}        &   2      \\ \hline

\begin{tabular}[c]{@{}l@{}}Number of\\ Multi-Heads\end{tabular}& \multicolumn{1}{c|}{ \begin{tabular}[c]{@{}c@{}}Categorical\\ {[}1, 2, 4, 8{]} \end{tabular}  }             & \multicolumn{1}{c|}{4}        & \multicolumn{1}{c|}{8} & \multicolumn{1}{c|}{4}      & \multicolumn{1}{c|}{8}     & \multicolumn{1}{c|}{1}        & \multicolumn{1}{c|}{4}        & \multicolumn{1}{c|}{1}        &     8    \\ \hline

Dropout Rate                                 & \multicolumn{1}{c|}{\begin{tabular}[c]{@{}c@{}}Categorical\\ {[}0.0, 0.1, 0.2, 0.3, 0.4, 0.5{]}\end{tabular}}             & \multicolumn{1}{c|}{0.0}    & \multicolumn{1}{c|}{0.0}      & \multicolumn{1}{c|}{0.0}    & \multicolumn{1}{c|}{0.0}      & \multicolumn{1}{c|}{0.0}        & \multicolumn{1}{c|}{0.0}        & \multicolumn{1}{c|}{0.0}        &    0.0     \\ \hline

\begin{tabular}[c]{@{}l@{}}Embedding\\ Dimensions\end{tabular} & \multicolumn{1}{c|}{\begin{tabular}[c]{@{}c@{}}Categorical\\ {[}32, 64, 128, 256, 512, 1024{]} \end{tabular}}             & \multicolumn{1}{c|}{128}        & \multicolumn{1}{c|}{128} & \multicolumn{1}{c|}{128}  & \multicolumn{1}{c|}{128}      & \multicolumn{1}{c|}{64}        & \multicolumn{1}{c|}{32}        & \multicolumn{1}{c|}{128}        &     64    \\ \hline

\begin{tabular}[c]{@{}l@{}}Hidden\\ Dimension\end{tabular} & \multicolumn{1}{c|}{\begin{tabular}[c]{@{}c@{}}Categorical\\ {[}64, 128, 256, 512, 1024{]} \end{tabular}}             & \multicolumn{1}{c|}{512}        & \multicolumn{1}{c|}{512}    & \multicolumn{1}{c|}{512}      & \multicolumn{1}{c|}{512}      & \multicolumn{1}{c|}{128}        & \multicolumn{1}{c|}{256}        & \multicolumn{1}{c|}{128}        &      256   \\ \hline

\end{tabular}
\end{table*}


\begin{table*}[ht]
\renewcommand{\arraystretch}{1.20}
\centering
\caption{Results of all trained networks over all 10 folds for all input combinations. The value in parentheses represents the standard deviation. The metrics are calculated over all channels and over all electrodes. MAE is calculated on the output values in the original scale. Norm. MAE is calculated on the normalized output of the network; the lower, the better.}\label{tab:table_Results_2}
\setlength\tabcolsep{4pt} 

\begin{minipage}[t]{.0\textwidth}

\begin{tabular}{|l|c|c|c|c|c|c|}
\hline
\multicolumn{2}{|c|}{\multirow{2}{*}{}}                                                                                  & \multirow{2}{*}{\begin{tabular}[c]{@{}c@{}}Nb.\\ Param.\end{tabular}} & \multicolumn{1}{c|}{MAE}                                                     & \multicolumn{1}{c|}{\begin{tabular}[c]{@{}c@{}}Norm.\\ MAE\end{tabular}} & \multicolumn{1}{c|}{MAE}                                                     & \begin{tabular}[c]{@{}c@{}}Norm.\\ MAE\end{tabular}     \\ \cline{4-7} 
\multicolumn{2}{|c|}{}                                                                                                   &                                                                       & \multicolumn{2}{l|}{Over all Channels}                                                                                                                  & \multicolumn{2}{l|}{Electrodes Only}                                                                                                   \\ \hline

\multicolumn{1}{|l|}{\multirow{16}{*}{\rotatebox[origin=c]{90}{\begin{tabular}[c]{@{}c@{}}Ruppel et al.\cite{Ruppel2018}\end{tabular}}}}&  \multicolumn{1}{|l|}{\multirow{2}{*}{1}} & \multicolumn{1}{c|}{\multirow{2}{*}{806K}} & 18.977 & 0.228 & 17.147 & 0.232 \\
& & &(1.719) & (0.022) & (1.642) & (0.023) \\\cline{2-7}
& \multicolumn{1}{|l|}{\multirow{2}{*}{2}} & \multicolumn{1}{c|}{\multirow{2}{*}{805K}} & 23.220 & 0.237 & 18.941 & 0.239 \\
& & &(1.403) & (0.018) & (1.458) & (0.020) \\\cline{2-7}
& \multicolumn{1}{|l|}{\multirow{2}{*}{3}} & \multicolumn{1}{c|}{\multirow{2}{*}{804K}} & 25.739 & 0.245 & 19.012 & 0.240 \\
& & &(1.539) & (0.019) & (1.498) & (0.021) \\\cline{2-7}
& \multicolumn{1}{|l|}{\multirow{2}{*}{4}} & \multicolumn{1}{c|}{\multirow{2}{*}{807K}} & 21.944 & 0.233 & 19.003 & 0.240 \\
& & &(1.514) & (0.019) & (1.473) & (0.020) \\\cline{2-7}
& \multicolumn{1}{|l|}{\multirow{2}{*}{5}} & \multicolumn{1}{c|}{\multirow{2}{*}{819K}} & 21.809 & 0.234 & 19.194 & 0.242 \\
& & &(1.548) & (0.020) & (1.529) & (0.021) \\\cline{2-7}
& \multicolumn{1}{|l|}{\multirow{2}{*}{6}} & \multicolumn{1}{c|}{\multirow{2}{*}{812K}} & 22.319 & 0.234 & 19.038 & 0.240 \\
& & &(1.469) & (0.019) & (1.468) & (0.021) \\\cline{2-7}
& \multicolumn{1}{|l|}{\multirow{2}{*}{7}} & \multicolumn{1}{c|}{\multirow{2}{*}{809K}} & 22.497 & 0.234 & 18.811 & 0.238 \\
& & &(1.466) & (0.019) & (1.449) & (0.020) \\\cline{2-7}
& \multicolumn{1}{|l|}{\multirow{2}{*}{8}} & \multicolumn{1}{c|}{\multirow{2}{*}{822K}} & 21.555 & 0.231 & 18.927 & 0.239 \\
& & &(1.504) & (0.019) & (1.489) & (0.020) \\\hline

\multicolumn{1}{|l|}{\multirow{16}{*}{\rotatebox[origin=c]{90}{\begin{tabular}[c]{@{}c@{}}Our XGBoost\\ Regressor\end{tabular}}}} & \multicolumn{1}{|l|}{\multirow{2}{*}{1}} & \multicolumn{1}{c|}{\multirow{2}{*}{1584K}} & 13.368 & 0.150 & 11.446 & \textbf{0.150} \\
& & &(1.340) & (0.015) & (1.204) & (0.015) \\\cline{2-7}
& \multicolumn{1}{|l|}{\multirow{2}{*}{2}} & \multicolumn{1}{c|}{\multirow{2}{*}{1041K}} & 14.372 & 0.159 & 12.002 & 0.158 \\
& & &(1.446) & (0.016) & (1.298) & (0.016) \\\cline{2-7}
& \multicolumn{1}{|l|}{\multirow{2}{*}{3}} & \multicolumn{1}{c|}{\multirow{2}{*}{1209K}} & 15.994 & 0.168 & 12.373 & 0.163 \\
& & &(1.622) & (0.017) & (1.340) & (0.017) \\\cline{2-7}
& \multicolumn{1}{|l|}{\multirow{2}{*}{4}} & \multicolumn{1}{c|}{\multirow{2}{*}{1694K}} & 13.275 & 0.152 & 11.638 & 0.153 \\
& & &(1.430) & (0.016) & (1.335) & (0.016) \\\cline{2-7}
& \multicolumn{1}{|l|}{\multirow{2}{*}{5}} & \multicolumn{1}{c|}{\multirow{2}{*}{840K}} & 13.407 & 0.153 & 11.774 & 0.154 \\
& & &(1.393) & (0.015) & (1.277) & (0.016) \\\cline{2-7}
& \multicolumn{1}{|l|}{\multirow{2}{*}{6}} & \multicolumn{1}{c|}{\multirow{2}{*}{1046K}} & 13.785 & 0.156 & 11.866 & 0.156 \\
& & &(1.385) & (0.015) & (1.277) & (0.016) \\\cline{2-7}
& \multicolumn{1}{|l|}{\multirow{2}{*}{7}} & \multicolumn{1}{c|}{\multirow{2}{*}{1239K}} & 13.299 & 0.150 & 11.375 & \textbf{0.150} \\
& & &(1.330) & (0.015) & (1.182) & (0.015) \\\cline{2-7}
& \multicolumn{1}{|l|}{\multirow{2}{*}{8}} & \multicolumn{1}{c|}{\multirow{2}{*}{1331K}} & \textbf{12.873} & \textbf{0.148} & 11.385 & \textbf{0.150} \\
& & &(1.255) & (0.014) & (1.156) & (0.014) \\\hline

\end{tabular}

\end{minipage} \hfill
\begin{minipage}[t]{.4\textwidth}
\begin{tabular}{|l|c|c|c|c|c|c|}
\hline
\multicolumn{2}{|c|}{\multirow{2}{*}{}}                                                                                  & \multirow{2}{*}{\begin{tabular}[c]{@{}c@{}}Nb.\\ Param.\end{tabular}} & \multicolumn{1}{c|}{MAE}                                                     & \multicolumn{1}{c|}{\begin{tabular}[c]{@{}c@{}}Norm.\\ MAE\end{tabular}} & \multicolumn{1}{c|}{MAE}                                                     & \begin{tabular}[c]{@{}c@{}}Norm.\\ MAE\end{tabular}     \\ \cline{4-7} 
\multicolumn{2}{|c|}{}                                                                                                   &                                                                       & \multicolumn{2}{l|}{Over all Channels}                                                                                                                  & \multicolumn{2}{l|}{Electrodes Only}                                                                                                   \\ \hline


\multicolumn{1}{|l|}{\multirow{16}{*}{\rotatebox[origin=c]{90}{\begin{tabular}[c]{@{}c@{}}Our Feed-Forward\\ Neural Network\end{tabular}}}} & \multicolumn{1}{|l|}{\multirow{2}{*}{1}} & \multicolumn{1}{c|}{\multirow{2}{*}{2233K}} & 14.693 & 0.168 & 12.724 & 0.169 \\
& & &(1.386) & (0.015) & (1.252) & (0.016) \\\cline{2-7}
& \multicolumn{1}{|l|}{\multirow{2}{*}{2}} & \multicolumn{1}{c|}{\multirow{2}{*}{2881K}} & 15.624 & 0.173 & 12.965 & 0.171 \\
& & &(1.536) & (0.017) & (1.388) & (0.018) \\\cline{2-7}
& \multicolumn{1}{|l|}{\multirow{2}{*}{3}} & \multicolumn{1}{c|}{\multirow{2}{*}{1701K}} & 17.010 & 0.181 & 13.265 & 0.175 \\
& & &(1.314) & (0.014) & (1.105) & (0.014) \\\cline{2-7}
& \multicolumn{1}{|l|}{\multirow{2}{*}{4}} & \multicolumn{1}{c|}{\multirow{2}{*}{1478K}} & 14.712 & 0.171 & 13.109 & 0.173 \\
& & &(1.317) & (0.015) & (1.247) & (0.016) \\\cline{2-7}
& \multicolumn{1}{|l|}{\multirow{2}{*}{5}} & \multicolumn{1}{c|}{\multirow{2}{*}{2554K}} & 14.056 & 0.166 & 12.721 & 0.169 \\
& & &(1.213) & (0.013) & (1.100) & (0.014) \\\cline{2-7}
& \multicolumn{1}{|l|}{\multirow{2}{*}{6}} & \multicolumn{1}{c|}{\multirow{2}{*}{1124K}} & 14.476 & 0.167 & 12.676 & 0.168 \\
& & &(1.150) & (0.013) & (1.043) & (0.013) \\\cline{2-7}
& \multicolumn{1}{|l|}{\multirow{2}{*}{7}} & \multicolumn{1}{c|}{\multirow{2}{*}{1794K}} & 15.231 & 0.177 & 13.351 & 0.178 \\
& & &(1.344) & (0.016) & (1.242) & (0.016) \\\cline{2-7}
& \multicolumn{1}{|l|}{\multirow{2}{*}{8}} & \multicolumn{1}{c|}{\multirow{2}{*}{2490K}} & 14.517 & 0.172 & 13.204 & 0.175 \\
& & &(1.450) & (0.017) & (1.404) & (0.017) \\\hline

\multicolumn{1}{|l|}{\multirow{16}{*}{\rotatebox[origin=c]{90}{\begin{tabular}[c]{@{}c@{}}Our Transformer\\ Encoder\end{tabular}}}} & \multicolumn{1}{|l|}{\multirow{2}{*}{1}} & \multicolumn{1}{c|}{\multirow{2}{*}{599K}} & 13.760 & 0.156 & 11.736 & 0.155 \\
& & &(1.400) & (0.016) & (1.280) & (0.016) \\\cline{2-7}
& \multicolumn{1}{|l|}{\multirow{2}{*}{2}} & \multicolumn{1}{c|}{\multirow{2}{*}{203K}} & 14.534 & 0.162 & 12.102 & 0.160 \\
& & &(1.482) & (0.017) & (1.374) & (0.017) \\\cline{2-7}
& \multicolumn{1}{|l|}{\multirow{2}{*}{3}} & \multicolumn{1}{c|}{\multirow{2}{*}{598K}} & 15.229 & 0.160 & 11.595 & 0.154 \\
& & &(1.617) & (0.017) & (1.328) & (0.017) \\\cline{2-7}
& \multicolumn{1}{|l|}{\multirow{2}{*}{4}} & \multicolumn{1}{c|}{\multirow{2}{*}{401K}} & 13.204 & 0.152 & 11.564 & 0.153 \\
& & &(1.363) & (0.016) & (1.266) & (0.016) \\\cline{2-7}
& \multicolumn{1}{|l|}{\multirow{2}{*}{5}} & \multicolumn{1}{c|}{\multirow{2}{*}{237K}} & 13.441 & 0.155 & 11.824 & 0.156 \\
& & &(1.241) & (0.014) & (1.147) & (0.015) \\\cline{2-7}
& \multicolumn{1}{|l|}{\multirow{2}{*}{6}} & \multicolumn{1}{c|}{\multirow{2}{*}{114K}} & 13.669 & 0.155 & 11.719 & 0.155 \\
& & &(1.505) & (0.018) & (1.376) & (0.018) \\\cline{2-7}
& \multicolumn{1}{|l|}{\multirow{2}{*}{7}} & \multicolumn{1}{c|}{\multirow{2}{*}{701K}} & 13.927 & 0.158 & 11.838 & 0.157 \\
& & &(1.614) & (0.019) & (1.468) & (0.019) \\\cline{2-7}
& \multicolumn{1}{|l|}{\multirow{2}{*}{8}} & \multicolumn{1}{c|}{\multirow{2}{*}{\textbf{103K}}} & 12.984 & 0.149 & \textbf{11.334} & \textbf{0.150} \\
& & &(1.552) & (0.018) & (1.473) & (0.019) \\ \hline

\end{tabular}
\end{minipage}

\end{table*}


\begin{table*}[ht]
\renewcommand{\arraystretch}{1.20}
\centering
\caption{Extended Significance Test with the corrected paired $t$-test~\cite{Nadeau_2003_inference} conducted on different input combinations for all networks.  The first value depicts the paired normalized MAE difference in percent over the ten folds,
the second value represents $t$-statistic, and the third value between parenthesis represents the $p$-value.}\label{tab:t_test_extended_1}
\setlength\tabcolsep{4pt}
\begin{tabular}{|l|c|c|c|c|c|c|c|c|c|}
\hline
 & 1 vs 2 & 1 vs 3 & 1 vs 4 & 1 vs 5 & 1 vs 6 & 7 vs 1 & 8 vs 1 & 5 vs 6 & 8 vs 5   \\ \hline
\multicolumn{1}{|l|}{\multirow{3}{*}{Ruppel et al.~\cite{Ruppel2018}}}     & -0.893\% & -1.763\% & -0.513\% & -0.630\% & -0.655\% & 0.578\% & 0.353\% & -0.025\% & -0.277\% \\
 & \textbf{(0.004)} & \textbf{(0.000)} & \textbf{(0.044)} & \textbf{(0.009)} & \textbf{(0.013)} & (0.967) & (0.915) & (0.398) & \textbf{(0.007)} \\\hline

Our XGBoost    & -0.859\% & -1.773\% & -0.113\% & -0.282\% & -0.539\% & -0.055\% & -0.208\% & -0.258\% & -0.490\% \\

 Regressor  & \textbf{(0.000)} & \textbf{(0.000)} & (0.129) & \textbf{(0.001)} & \textbf{(0.000)} & (0.190) & \textbf{(0.009)} & \textbf{(0.009)} & \textbf{(0.001)} \\\hline

Our Feed-Forward  & -0.441\% & -1.267\% & -0.324\% & 0.217\% & 0.133\% & 0.862\% & 0.402\% & -0.084\% & 0.620\% \\
Neural Network & -2.222 & -5.436 & -2.184 & 0.589 & 0.373 & 1.995 & 0.729 & -0.226 & 1.296 \\
 & \textbf{(0.027)} & \textbf{(0.000)} & \textbf{(0.028)} & (0.715) & (0.641) & (0.961) & (0.758) & (0.413) & (0.886) \\\hline

Our Transformer  & -0.587\%& -0.391\% & 0.385\% & 0.077\% & 0.083\% & 0.188\% & -0.662\%& 0.007\%& -0.585\% \\
Encoder  & -2.479 & -1.400 & 1.074 & 0.344 & 0.438 & 0.461 & -1.870 & 0.019 & -1.537 \\
 & \textbf{(0.018)} & (0.097) & (0.845) & (0.631) & (0.664) & (0.672) & \textbf{(0.047)} & (0.507) & (0.079) \\\hline

\end{tabular}
\end{table*}


\begin{table*}[ht]
\renewcommand{\arraystretch}{1.20}
\centering
\caption{Extended Significance Test with the corrected paired $t$-test~\cite{Nadeau_2003_inference} conducted for all network pairs. The first value depicts the paired normalized MAE difference in percent over the ten folds,
the second value represents $t$-statistic, and the third value between parenthesis represents the $p$-value.}\label{tab:t_test_extended_2}
\setlength\tabcolsep{4pt}
\begin{tabular}{|l|c|c|c|c|c|c|}
\hline
\multicolumn{1}{|l|}{\multirow{2}{*}{vs}}&  Our XGBoost & Our FFNN &  Our Transformer& Our XGBoost & Our XGBoost &  Our Transformer   \\
& Ruppel et al.~\cite{Ruppel2018} & Ruppel et al.~\cite{Ruppel2018}   & Ruppel et al.~\cite{Ruppel2018}  &  Our FFNN &  Our Transformer  & Our FFNN  \\ \hline

\multicolumn{1}{|l|}{\multirow{3}{*}{1}} 
 & -7.740\% & -5.962\% & -7.209\% & -1.778\% & -0.531\% & -1.247\% \\
 & -9.785 & -7.741 & -8.670 & -5.188 & -1.581 & -3.690 \\
 & \textbf{(0.000)} & \textbf{(0.000)} & \textbf{(0.000)} & \textbf{(0.000)} & (0.074) & \textbf{(0.002)} \\\hline
\multicolumn{1}{|l|}{\multirow{3}{*}{2}} 
 & -7.774\% & -6.414\% & -7.515\% & -1.360\% & -0.259\% & -1.101\% \\
 & -9.962 & -8.136 & -8.956 & -4.284 & -0.704 & -5.338 \\
 & \textbf{(0.000)} & \textbf{(0.000)} & \textbf{(0.000)} & \textbf{(0.001)} & (0.250) & \textbf{(0.000)} \\\hline
\multicolumn{1}{|l|}{\multirow{3}{*}{3}} 
 & -7.730\% & -6.457\% & -8.581\% & -1.272\% & 0.851\% & -2.123\% \\
 & -9.936 & -8.509 & -11.401 & -3.623 & 3.407 & -6.570 \\
 & \textbf{(0.000)} & \textbf{(0.000)} & \textbf{(0.000)} & \textbf{(0.003)} & (0.996) & \textbf{(0.000)} \\\hline
\multicolumn{1}{|l|}{\multirow{3}{*}{4}} 
 & -8.140\% & -6.151\% & -8.108\% & -1.989\% & -0.032\% & -1.956\% \\
 & -9.546 & -7.352 & -8.958 & -5.653 & -0.106 & -11.260 \\
 & \textbf{(0.000)} & \textbf{(0.000)} & \textbf{(0.000)} & \textbf{(0.000)} & (0.459) & \textbf{(0.000)} \\\hline
\multicolumn{1}{|l|}{\multirow{3}{*}{5}} 
 & -8.089\% & -6.809\% & -7.916\% & -1.279\% & -0.173\% & -1.107\% \\
 & -9.960 & -8.364 & -10.944 & -3.888 & -0.477 & -3.810 \\
 & \textbf{(0.000)} & \textbf{(0.000)} & \textbf{(0.000)} & \textbf{(0.002)} & (0.322) & \textbf{(0.002)} \\\hline
\multicolumn{1}{|l|}{\multirow{3}{*}{6}} 
 & -7.856\% & -6.750\% & -7.947\% & -1.106\% & 0.092\% & -1.197\% \\
 & -10.032 & -7.725 & -9.859 & -3.610 & 0.303 & -3.918 \\
 & \textbf{(0.000)} & \textbf{(0.000)} & \textbf{(0.000)} & \textbf{(0.003)} & (0.616) & \textbf{(0.002)} \\\hline
\multicolumn{1}{|l|}{\multirow{3}{*}{7}} 
 & -8.373\% & -5.678\% & -7.599\% & -2.696\% & -0.774\% & -1.921\% \\
 & -10.568 & -7.579 & -7.886 & -5.235 & -1.347 & -5.899 \\
 & \textbf{(0.000)} & \textbf{(0.000)} & \textbf{(0.000)} & \textbf{(0.000)} & (0.105) & \textbf{(0.000)} \\\hline
\multicolumn{1}{|l|}{\multirow{3}{*}{8}} 
 & -8.301\% & -5.912\% & -8.224\% & -2.389\% & -0.077\% & -2.312\% \\
 & -10.363 & -5.268 & -9.773 & -4.242 & -0.157 & -4.004 \\
 & \textbf{(0.000)} & \textbf{(0.000)} & \textbf{(0.000)} & \textbf{(0.001)} & (0.439) & \textbf{(0.002)} \\\hline

\end{tabular}
\end{table*}

\begin{table*}[ht]
\renewcommand{\arraystretch}{1.20}
\centering
\caption{Number of parameters in thousands, inference time in milliseconds, and floating-point operations per second (FLOPS) in millions for all approaches across all input combinations. The number of FLOPS is only calculated for the neural networks.}\label{tab:param_inference_flops}
\setlength\tabcolsep{6pt}
\begin{tabular}{|l|c|c|c|c|c|c|c|c|c|c|c|c|}
\hline
 & \multicolumn{3}{|c|}{Ruppel et al.'s Network $B$~\cite{Ruppel2018}}  & \multicolumn{3}{c|}{Our XGBoost Regressor}  & \multicolumn{3}{c|}{Our FF Neural Network}  &   \multicolumn{3}{c|}{Our Transformer Encoder}   \\ \cline{2-13} 
& Numb. & Inference &  Numb. & Numb. & Inference&  Numb. & Numb.& Inference&   Numb. &  Numb. & Inference &  Numb.  \\ 

& Param. &  ($ms$) &FLOPS &Param. & ($ms$)& FLOPS &Param. & ($ms$)& FLOPS &Param. & ($ms$)& FLOPS  \\ \hline

1 & 806K & 0.628 & 1.61M & 1584K & 0.422 & -  & 2233K & 0.876 & 4.46M & 599K & 1.618 & 5.97M \\\hline
2 & 805K & 0.539 & 1.61M & 1041K & 0.272 & -  & 2881K & 1.173 & 5.75M & 203K & 2.982 & 1.63M \\\hline
3 & 804K & 0.539 & 1.61M & 1209K & 0.358 & -  & 1701K & 0.741 & 3.40M & 598K & 1.396 & 3.58M \\\hline
4 & 807K & 0.555 & 1.61M & 1694K & 0.432 & -  & 1478K & 0.757 & 2.95M & 401K & 1.005 & 5.59M \\\hline
5 & 819K & 0.572 & 1.65M & 840K & 0.304 & -  & 2554K & 0.833 & 5.10M & 237K & 2.995 & 11.64M \\\hline
6 & 812K & 0.552 & 1.62M & 1046K & 0.356 & -  & 1124K & 0.613 & 2.24M & 114K & 1.427 & 2.98M \\\hline
7 & 809K & 0.572 & 1.62M & 1239K & 0.387 & -  & 1794K & 0.760 & 3.58M & 701K & 2.939 & 9.89M \\\hline
8 & 822K & 0.607 & 1.65M & 1331K & 0.412 & -  & 2490K & 1.002 & 4.98M & 103K & 0.988 & 5.30M \\\hline
                        
\end{tabular}
\end{table*}

\end{document}